\pdfoutput=1 

\documentclass[twocolumn]{article}
\usepackage{adjustbox}
\usepackage{authblk}
\usepackage{amsmath}
\usepackage{amssymb} 
\usepackage{amstext}
\usepackage{booktabs}
\usepackage{caption}
\usepackage[greek, british]{babel}
\usepackage[utf8]{inputenc}
\usepackage{csquotes} 
\usepackage{makecell}
\usepackage{multirow}
\usepackage{textcomp}
\usepackage{tikz}
\usepackage{setspace}
\usepackage{siunitx}
\usepackage[caption=false]{subfig}
\usepackage{varioref}
\usetikzlibrary{intersections,3d,decorations.text,shapes.arrows,positioning,fit,backgrounds,trees,positioning,fit,arrows,decorations.pathreplacing, calc,math,through,arrows.meta }
\definecolor{tikzBoxColourTruePositive}{RGB}{90,105,0}
\definecolor{tikzBoxColourFalsePositive}{RGB}{214,122,62}
\definecolor{tikzBoxColourFalseNegative}{RGB}{117,147,176}
\newcommand{\tikzBoxSize}{0.15cm}
\DeclareRobustCommand\tikzBoxTruePositive{\tikz\node[rectangle,fill=tikzBoxColourTruePositive,minimum width=\tikzBoxSize,minimum height = \tikzBoxSize,] (r) at (0,0) {};}
\DeclareRobustCommand\tikzBoxFalsePositive{\tikz\node[rectangle,fill=tikzBoxColourFalsePositive,minimum width=\tikzBoxSize,minimum height = \tikzBoxSize,] (r) at (0,0) {};}
\DeclareRobustCommand\tikzBoxFalseNegative{\tikz\node[rectangle,fill=tikzBoxColourFalseNegative,minimum width=\tikzBoxSize,minimum height = \tikzBoxSize,] (r) at (0,0) {};}

\definecolor{fhg}{RGB}{23,156,125}
\usepackage[colorlinks=true, pdfstartview=FitV, allcolors=fhg, pagebackref=true]{hyperref}
\usepackage{cleveref} 

\DeclareRobustCommand\displayNameEngster{\emph{Team One}}
\DeclareRobustCommand\displayNameMichen{\emph{Team Two}}

\usepackage[colorinlistoftodos]{todonotes}
\usepackage{comment}

\title{Instance Segmentation XXL-CT Challenge of a Historic Airplane}
\author{
	\href{https://orcid.org/0000-0002-3429-732X}{\includegraphics[scale=0.06]{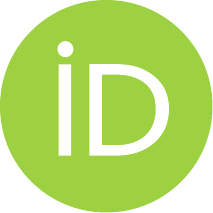}}Roland Gruber$^{1,2}$ 
	Johann Christopher Engster$^{3,4}$ 
	Markus Michen$^{1}$ \\
	Nele Blum$^{3,4}$ 
	Maik Stille$^{3,4}$ 
	\href{https://orcid.org/0000-0002-5590-2920}{\includegraphics[scale=0.06]{./data/full/orcid.pdf}}Stefan Gerth$^{1}$
	\href{https://orcid.org/0000-0003-0840-8695}{\includegraphics[scale=0.06]{./data/full/orcid.pdf}}Thomas Wittenberg$^{1,2}$\\ 
	
	$^{1}$ \small Fraunhofer IIS, Fraunhofer Institute for Integrated Circuits IIS, Fürth, Germany \\
	$^{2}$ \small Friedrich-Alexander-Universität Erlangen-Nürnberg, Germany \\
	$^{3}$ \small Fraunhofer Research Institution for Individualized and Cell-Based Medical Engineering (IMTE), Lübeck , Germany \\
	$^{4}$ \small Institute of Medical Engineering, University of Lübeck, Lübeck, Germany		
}

\date{2024-02-05}

\begin{document}
	\maketitle
	
	\begin{abstract}
		Instance segmentation of compound objects in XXL-CT imagery poses a unique challenge in non-destructive testing. This complexity arises from the lack of known reference segmentation labels, limited applicable segmentation  tools, as well as partially degraded image quality. To asses recent advancements in the field of machine learning-based image segmentation, the \enquote{Instance Segmentation XXL-CT Challenge of a Historic Airplane} was conducted. The challenge aimed to explore automatic or interactive instance segmentation methods for an efficient delineation of the different aircraft components, such as screws, rivets, metal sheets or pressure tubes. We report the organization and outcome of this challenge and describe the capabilities and limitations of the submitted segmentation methods.

	\end{abstract}
	
	\section{Introduction} 
		The task of computing a complete instance segmentation of large-scale (XXL) industrial Computed Tomography (CT) imagery for compound objects -- such as vehicles, boats, shipping containers, or even airplanes -- presents a unique and complex problem in the field of non-destructive testing (NDT) \cite{Gruber2020, Gruber2022}. The underlying difficulty of this challenge relates to the \enquote{single-lot} nature of the data as there exists a lack of known reference segmentation labels as well as a lack of applicable algorithms to handle the yet unknown data. Furthermore, physical phenomena and technical limitations which degrade image quality and contrast of the scans make the entity delineation challenging. However, recent advancements in machine learning-based image segmentation methods hold considerable potential to address this complex task. 	

		Together with the ADA Lovelace Center of the Fraunhofer Institute for Integrated Circuits IIS, the Fraunhofer Development Center for X-ray Technologies (EZRT) and the German Society for Non-Destructive Testing (DGZFP) an international XXL-CT instance segmentation challenge has been initiated and proclaimed, in order to find an answer to the following question: \enquote{Which automatic or interactive methods from the areas of digital image processing, machine learning or deep Neural networks can segment individual parts of a historic airplane with the highest quality?}	
		
		This challenge was designed as a technical competition to test and compare advanced algorithms for detecting and segmenting entities within XXL-CT data-sets. It specifically focuses on instance segmentation of data from a historic airplane. By evaluating different algorithms and comparing their performance, we intended to identify the most effective approaches for CT instance segmentation and provide insights into the strengths and weaknesses of current methods. Additionally, by connecting researchers in the field, we aimed to foster the collaboration between research institutes and facilitate the exchange of ideas and image analysis techniques.	Also by providing a standardized data-set and evaluation metrics, we create a benchmark to compare different instance segmentation algorithms and to identify areas for future research and improvement.
		
		The complete training data-set comprised seven sub-volume pairs with a dimension of $ 512 \times 512 \times 512$\,voxels extracted from a XXL-CT scan from a historic Me\,163 airplane. Each pair consisted of one reconstructed sub-volume and one corresponding annotated sub-volume, as detailed in \cite{Gruber2022}. The objective for the challenge participants was to develop adequate algorithms capable of accurately segmenting individual entities within an additional testing sub-volume $V_\text{test}$ of same dimensions. $V_\text{test}$ was extracted from the same XXL-CT scan and was located in close proximity to the training sub-volumes. Notably, the annotation of this test sub-volume was withheld and not provided to the participants (refer to Section \ref{sec:challenge-dataset}).
		
		The size and volume of industrial XXL-CT data-sets are often substantial, with numerous strongly inter-connected entities consisting of millions of voxels, making  manual segmentation of every single entity a time-consuming and error-prone task. To partially tackle this challenge, some methods have already been developed to approach instance segmentation of large-scale industrial CT volumes and address the issues at hand. For example, we recently proposed the use of \emph{flood filling networks} \cite{Gruber2020} which apply deep convolutional neural networks (DCNNs) embedded into a flood-filling algorithm. Ohtake et al. make use of region-growing approaches and the uniform thickness of sheet metals for the segmentation of XXL-CT scans of car bodies \cite{Ohtake2023}. The intent of these XXL instance segmentation approaches is to reduce the time and resources required for image analysis and segmentation while simultaneously improving the reliability of the achieved results.
			
		This paper follows the guidelines provided in \cite{63-21-MaierHein2018, 63-21, 63-21-AppendixA} for reporting the outcomes of challenges \cite{63}. 
	
	\section{Methods} 
		\subsection{Challenge Organization}	\label{sec:challenge-organization}	 
		
			Between December 2022 and March 2023, extensive preparations were made for the challenge logistics. These included the finalization of the seven manually annotated training data-sets derived from airplane CT Data (see Section \ref{sec:challenge-dataset}), the establishment of legal terms for data sharing and usage (together with the \enquote{Deutsches Museum} as the owner of the airplane), and the development and implementation of an online portal to facilitate the challenge proceedings.
			
			In January of 2023, the \emph{preparation} phase of the challenge started. We placed \enquote{calls for participation} to the interested image processing and NDT community through social media platforms such as LinkedIn and Reddit, as well as through \enquote{Papers With Code}, blog posts, and personalized emails. A call-for-participation was also included in the \enquote{ZfP-Zeitung} periodical of the DGZfP. By these calls for participation interested parties and individuals were invited to download a sample data-set to familiarize themselves with the data format and task at hand. Furthermore, we shared the information about the challenge during the poster sessions of two international conferences: the 12th International Conference on Industrial Computed Tomography (iCT2023) and the Conference on Optimization and Machine Learning in Industry (CONLI23). 		
			A preprint article \cite{Gruber2022} detailing the challenges and problems of the data-set was also released prior to the start of the challenge. During this \emph{preperation} phase interested parties were able to register via the challenge website and hence participate in the subsequent phases from that time onwards.					
			
			In mid-February 2023, the challenge transitioned to the \emph{training} phase. Registered participants were granted access to the training data-sets consisting out of seven training sub-volume pairs. Participants who registered after this period were granted immediate access to the same training data-sets.
			
		 	In May 2023, the \emph{testing} phase of the challenge was initiated, during which the test data-set, which had no accompanying reference labels, was provided to the participants. This test data-set consisted of an yet unseen sub-volume of the airplane and required the participants to perform a (semi-) automatic instance segmentation of the contained entities based on knowledge gained from the training data-set. The participants were given a one-week time frame to prepare and submit a proposed segmentation of the test data-set. This short time frame was chosen to minimize the possibility of excessive manual segmentation of the data-set. To simplify the challenge and reduce possible pressure on the participants, we requested only the proposed segmentation and a brief description of the applied algorithm, rather than asking for code or providing an interactive leader board. This approach intended to prevent participants from withholding seemingly underperforming segmentation results or refraining from submitting to the challenge as if they needed to release their own code. The goal of the challenge was not to identify the most accurate segmentation result, but to identify several valid methods addressing the problem at hand.
		 	
		 	The challenge concluded with the \emph{evaluation} and \emph{publication} phases starting May 2023. Due to late registrations and lacking submissions we extended the \emph{testing} phase of the challenge until the end of August 2023. At that point the challenge was closed.		
		 	
		 	Members of other departments of the organizing institute were also permitted to participate in the challenge but were not ranked. No awards were provided due to the public funding restrictions on the project. The top-performing participants are featured in a combined publication. After this publication, the participants are free to publish their detailed results by themself. Participants were asked to submit their results to the organizers via email. All instructions were sent to participants through email and a private forum.
		 	
		 	
		 	%

		\subsection{Challenge Mission} 
			The mission of the international XXL-CT instance segmentation challenge, was to uncover the most effective automatic or interactive instance and object segmentation methods in the fields of digital image processing, machine learning, and deep neural networks. The challenge aimed to identify techniques which can accurately and precisely delineate various components within a historic airplane. By addressing this mission, the challenge seeks to advance the understanding and application of the image segmentation process within the domain of NDT, ultimately contributing to the development of innovative solutions for complex object and instance segmentation tasks.
				
		\subsection{Challenge Dataset} \label{sec:challenge-dataset}
\begin{figure}
				\definecolor{tikzBoxColourDatasetTraining}{RGB}{23,156,125}
				\DeclareRobustCommand\tikzBoxDatasetTraining{\tikz\node[rectangle,fill=tikzBoxColourDatasetTraining,minimum width=\tikzBoxSize,minimum height = \tikzBoxSize,] (r) at (0,0) {};}
				
				\definecolor{tikzBoxColourDatasetTest}{RGB}{214,122,62}
				\DeclareRobustCommand\tikzBoxDatasetTest{\tikz\node[rectangle,fill=tikzBoxColourDatasetTest,minimum width=\tikzBoxSize,minimum height = \tikzBoxSize,] (r) at (0,0) {};}				
				
				\centering{}
				\includegraphics[width=0.8\columnwidth,height=1.0\textheight,keepaspectratio]{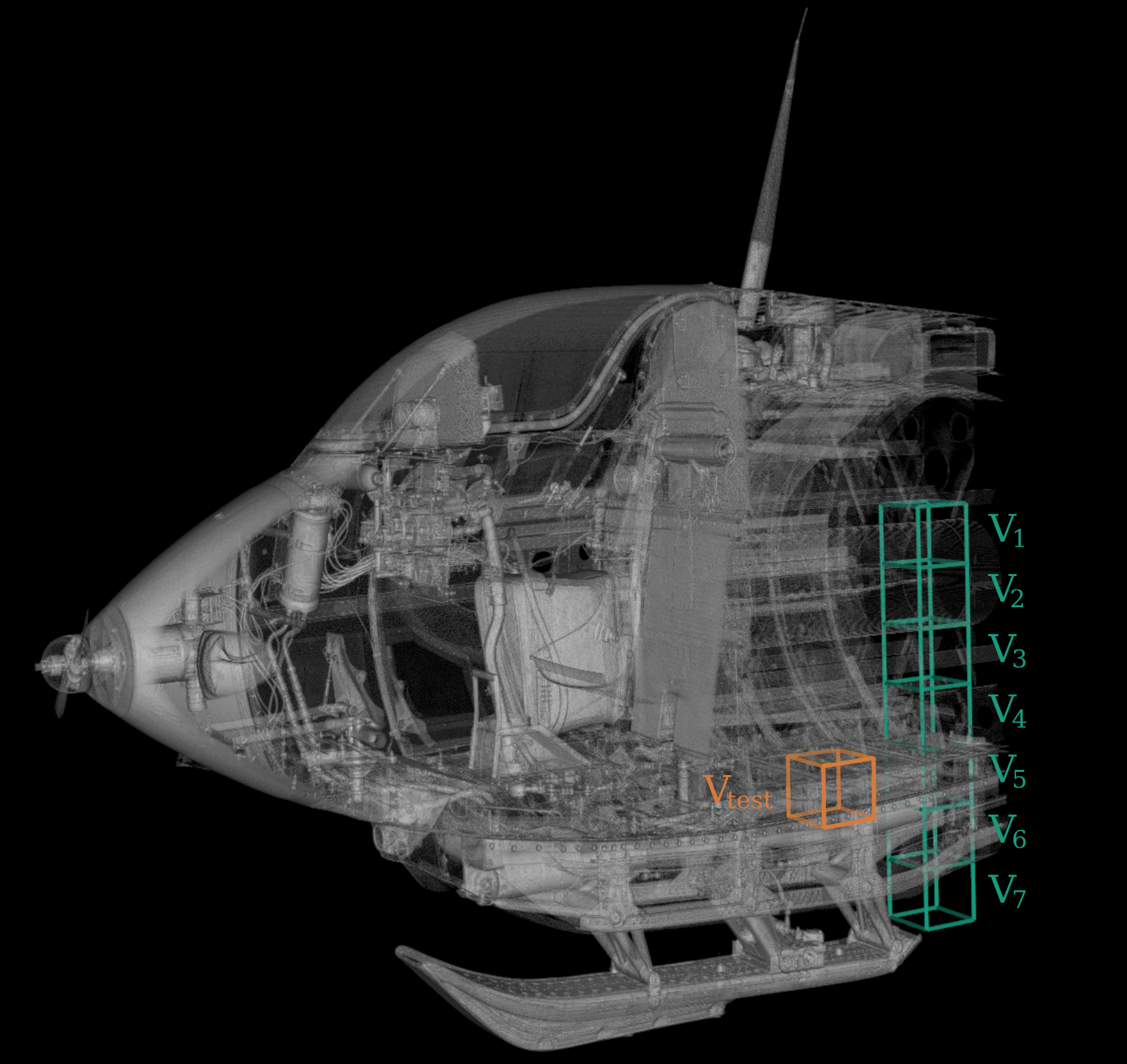}
				\caption{\label{fig:description-location}Clipped 3D-rendering of the XXL-CT reconstruction of the front part of the historical Me\,163 airplane containing the seven training sub-volumes $V_1, \dotsc, V_7$ (\tikzBoxDatasetTraining) as well as the sub-volume $V_{\text{test}}$  (\tikzBoxDatasetTest) used for the \emph{test} phase of the challenge.}
			\end{figure}

			While the training data set consisted of seven sub-volumes, containing both the CT-scan and corresponding reference segmentation, the testing data set was limited to one sub-volume, without accompanying reference labels. The relative positions of the testing and training sub-volumes within the XXL-CT reconstruction are illustrated in Figure \ref{fig:description-location}.
			All sub-volumes had dimensions of $512 \times 512 \times 512$ voxels. It is important to note that the testing sub-volume $V_\text{test}$ is not directly connected to any of the training sub-volumes, but it has been chosen to be comparable to training sub-volumes $V_5$ and $V_6$. 
			
			Table \ref{tab:description-segmentation} provides a comparison of key metrics between the testing and training data-sets, highlighting important differences between the sub-volumes, regarding the amount and size of segments, as well as the percentage of foreground voxels.
			
			The data-set acquisition and challenging aspects, such as noisy data, low contrast, and limited spatial resolution, have already been described in detail in resent works \cite{Gruber2022}. Furthermore the same work provides a comprehensive overview of the annotation and postprocessing process applied to gather the training and testing data.				
	
			\begin{table*}
				\begin{adjustbox}{max width=\textwidth}
					\begin{tabular}{llrrrrcr} \toprule
						Sub -  			& Origin 			& \# Segments		& Minimum 		& Maximum 		 & Median 			& Foreground\\ 
						volume 			& coordinate		& (prior post- 		& {segment size}& {segment size} & {segment size}  	& voxels  	\\ 
						                & of sub-volume		& processing)		& [voxels] 		& [voxels] 		& [voxels] 			& [\%]		\\ \midrule
						
						$V_1$ 			& (3072, 4608, 0)	& 5 (5) 			& 2,468 	& 1,210,717 		& 47,849 	& 1.3 		\\ 
						$V_2$ 			& (3072, 5120, 0) 	& 7 (7) 			& 1,277 	& 1,173,579			&\bf{78,659}& 1.1 		\\ 
						$V_3$ 			& (3072, 5632, 0) 	& 14 (13)			& 1,147 	& 889,731			& 9,790 	& 1.3 		\\ 
						$V_4$ 			& (3072, 6144, 0) 	& 33 (33) 			& 1,293		& \bf{1,768,078}	& 2,217		& 4.5 		\\ 
						$V_5$ 			& (3072, 6656, 0) 	& {\bf 169 (159)} 	& 187	 	& 1,545,773			& 3,853		& \bf{9.4}	\\ 
						$V_6$ 			& (3072, 7168, 0) 	& 108 (94)			& 158		& 1,567,273			& 4,039		& 5.6 		\\ 
						$V_7$ 			& (3072, 7680, 0) 	& 9 (9) 			& 509 		& 73,302			& 1,209		& 0.1 		\\
						$V_\text{test}$	& (3072, 6656, 1024)& 135 (134) 		& \bf{136} 	& 1,268,371			& 4,008		& 6.5 		\\ \bottomrule
					\end{tabular}
				\end{adjustbox}		
				\caption{\label{tab:description-segmentation} Key metrics of the annotated training sub-volumes $V_1$ to $V_7$ compared to the test sub-volume $V_\text{test}$.}
			\end{table*}
			
\begin{figure*}
				\begin{centering}
					\subfloat[Sub-volume \label{fig:description-example-volume}]					{\includegraphics[width=0.33\textwidth,height=0.25\textheight,keepaspectratio]						{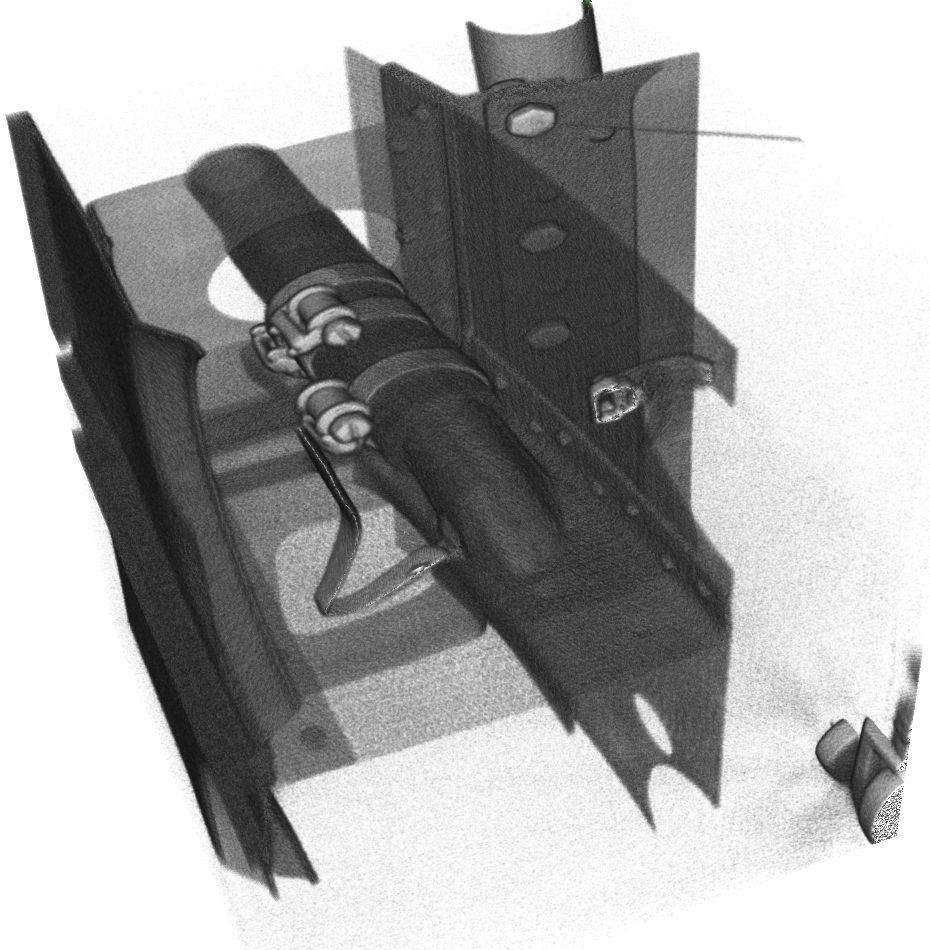}}	
					~\subfloat[Manually annotated reference \label{fig:description-example-all}]					{\includegraphics[width=0.33\textwidth,height=0.25\textheight,keepaspectratio]						{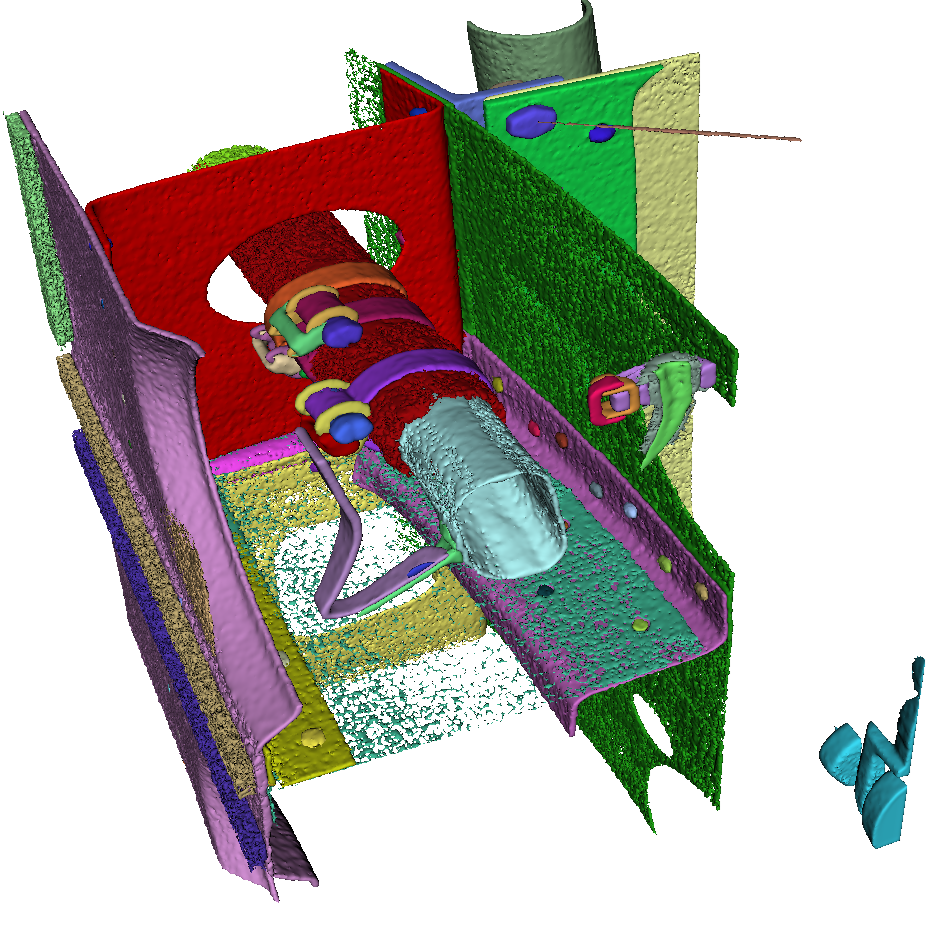}}
					~\subfloat[Metal sheets\label{fig:description-example-metallSheet}]					{\includegraphics[width=0.33\textwidth,height=0.25\textheight,keepaspectratio]						{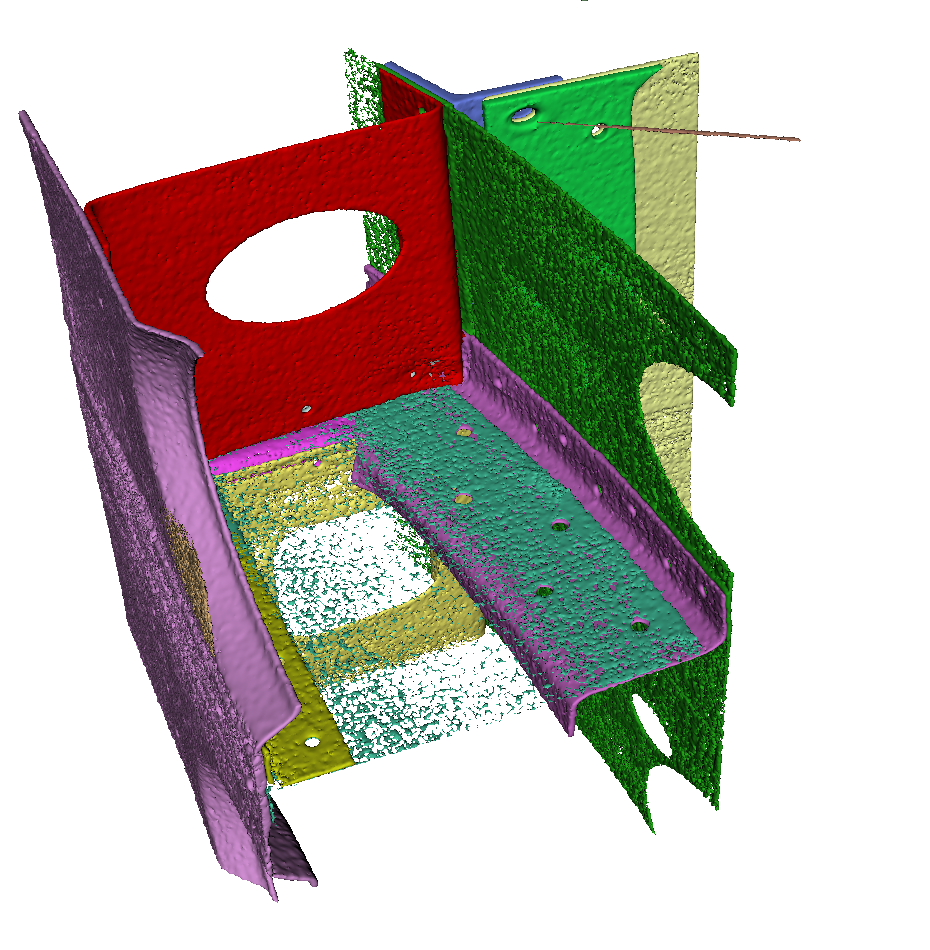}}
					
					\subfloat[Pipes \label{fig:description-example-pipes}]					{\includegraphics[width=0.33\textwidth,height=0.25\textheight,keepaspectratio]						{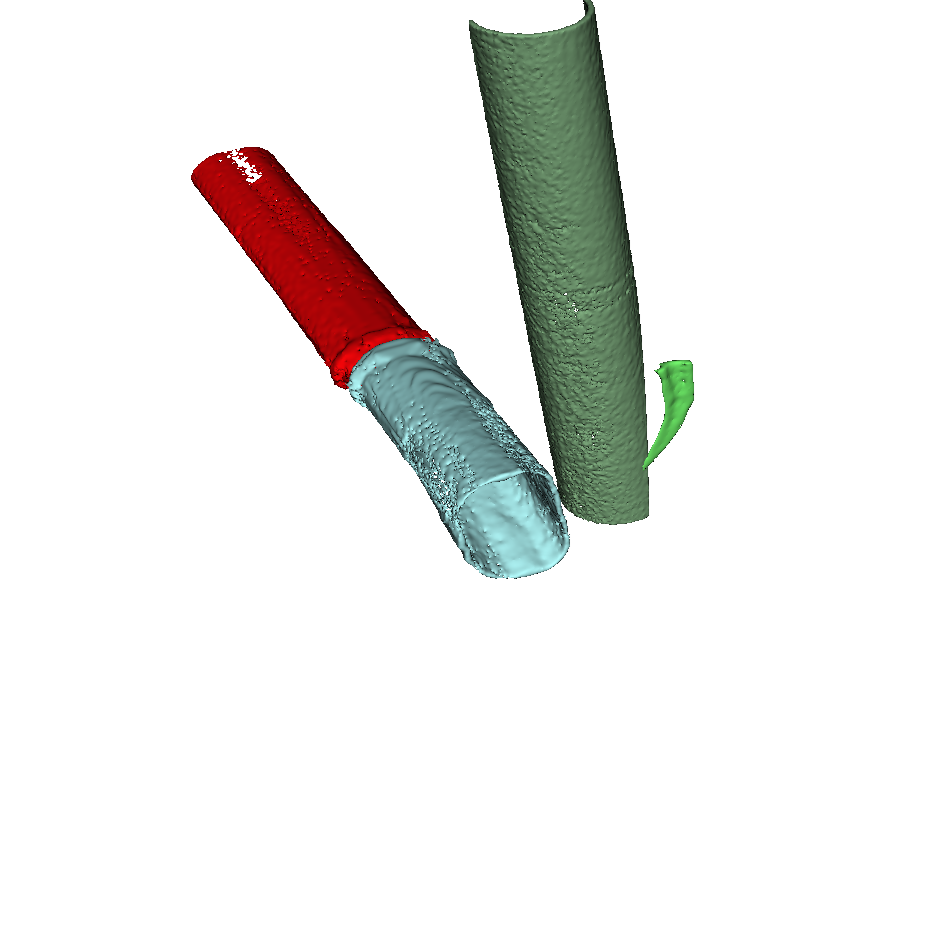}}	
					~\subfloat[Rivets and bolts \label{fig:description-example-rivetsAndBolts}]					{\includegraphics[width=0.33\textwidth,height=0.25\textheight,keepaspectratio]						{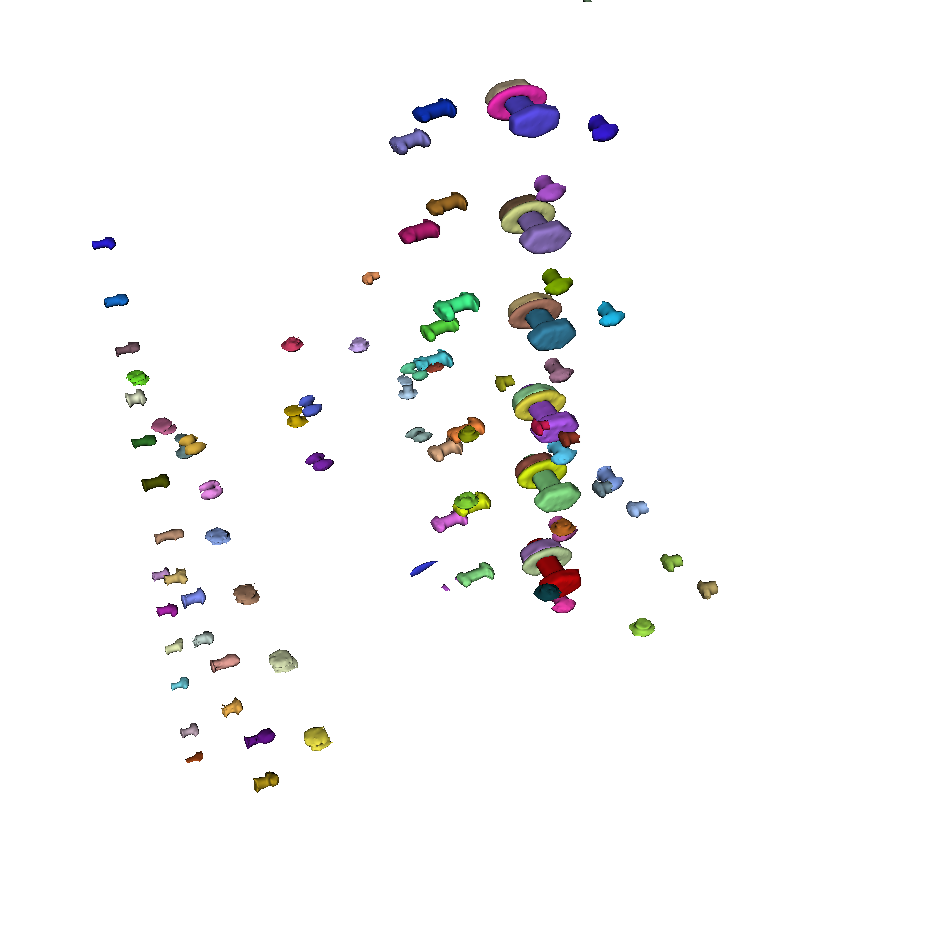}}	
					~\subfloat[Miscellaneous \label{fig:description-example-mountingAndMiscellaneous}]					{\includegraphics[width=0.33\textwidth,height=0.25\textheight,keepaspectratio]						{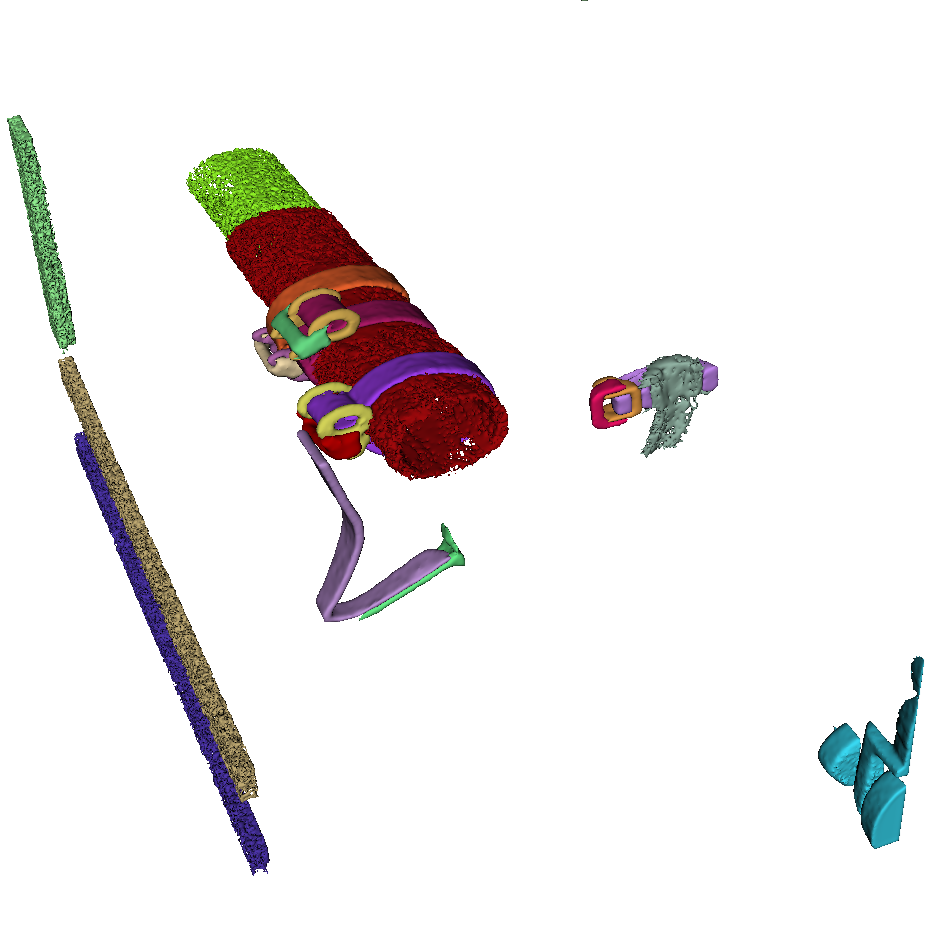}}
				\end{centering}
				\caption{\label{fig:description-example} Renderings of sub-volume $V_\text{test}$ (3072, 6656, 1024). While Figure \ref{fig:description-example-volume} shows the unannotated sub-volume, Figure \ref{fig:description-example-all} depicts all manually labelled segments separated by colour. To increase clarity, only the segments of a specific category are shown in the following sub-figures: Figure \ref{fig:description-example-metallSheet} shows all metal sheets; Figure \ref{fig:description-example-pipes} depicts the presumably pressure-carrying pipes, pressure tanks and lines; Figure \ref{fig:description-example-rivetsAndBolts} contains all rivets and screw connections; Figure \ref{fig:description-example-mountingAndMiscellaneous} shows all brackets, clamp connectors, and other miscellaneous transition elements that could not otherwise be assigned a category.}
			\end{figure*}

			Figure \ref{fig:description-example} showcases various renderings of the testing sub-volume $V_\text{test}$ (3072, 6656, 1024). Figure \ref{fig:description-example-volume} displays the unannotated volume, while Figure \ref{fig:description-example-all} presents all labelled segments in separate colours to distinguish them from one another. For clarity, only the segments of a particular category are shown in the following figures: Figure \ref{fig:description-example-metallSheet} exhibits all metal sheets, Figure \ref{fig:description-example-pipes} illustrates the presumed pressure-carrying pipes, pressure tanks, and lines, Figure \ref{fig:description-example-rivetsAndBolts} contains all rivets and screw connections, and finally, Figure \ref{fig:description-example-mountingAndMiscellaneous} displays all brackets, clamp connectors, and other miscellaneous transition elements that could not otherwise be assigned to a distinct category. The classes of individual segments were not relevant for the challenge. Furthermore, these classes cannot be accurately determined as many components do not fit into simple categories or fulfil multiple functions simultaneously.
			
		\subsection{Assessment method} \label{methods-AssesmentMethod}
			
			In our assessment approach, we make use of a \enquote{segment correlation matrix} to compare the manual reference segmentation of $V_\text{test}$ (3072, 6656, 1024) with the proposed segmentations. This matrix is constructed by assigning each row to a \enquote{reference segment} $S_R(i)$ and each column to a \enquote{detected segment} $S_D(j)$. The cells of the matrix correspond to the intersection over union (IoU) score, which is a measure of overlap between the two segments $S_R(i)$ and $S_D(j)$. A score of $1.0$ indicates complete overlap, while a score of $0.0$ indicates no overlap. All other values map to a range of $\text{IoU} \in [0, 1].$ 
			
			The rows of the matrix are sorted in descending order by the count of voxels of their corresponding reference segments. In other words, the largest segments are placed at the top, while the smallest segments are at the bottom. Similarly, the columns are sorted by searching for the detected segment with the best match or the highest IoU to the reference segment of the current row. Detected segments which do not match any reference segment are sorted by their voxel count. We excluded segments with a voxel count of fewer than 100 voxels to reduce the size of the matrix as the smallest reference segment had a voxel count of only 136\,voxels (see Table \ref{tab:description-segmentation}).
			
			In an ideal scenario, a \enquote{perfect segmentation} $S$ should be reflected by a quadratic correlation matrix containing the same count of rows and columns, and thus, the same amount of reference segments and detected segments. Moreover, all correlation values outside the main diagonal should contain IoU values of $0.0$, while all values on the main diagonal should have values of $1.0$. 
			
			However, in realistic applications, the row and column count may differ, and boundary errors could result in suboptimal correlation values. Rows with multiple horizontal values could denote an over-segmentation of the respective detected segment, or a reference segment which was accidentally split into multiple detected segments. In contrast, vertical lines indicate the merging of multiple reference segments into a single segment. Low values in the main diagonal indicate reference segments without a high IoU match in the detected segments. For further details and explanations of the assessment method, refer to \cite{Gruber2022}.
	
	\section{Results} 
	
		\subsection{Challenge Submission}

\begin{figure*}
				\definecolor{tikzBoxColourSubmitted}{RGB}{0,128,0}
				\DeclareRobustCommand\tikzBoxSubmitted{\tikz\node[rectangle,fill=tikzBoxColourSubmitted,minimum width=\tikzBoxSize,minimum height = \tikzBoxSize,] (r) at (0,0) {};}
				
				\definecolor{tikzBoxColourWithdrew}{RGB}{128,0,0}
				\DeclareRobustCommand\tikzBoxWithdrew{\tikz\node[rectangle,fill=tikzBoxColourWithdrew,minimum width=\tikzBoxSize,minimum height = \tikzBoxSize,] (r) at (0,0) {};}
				
				\definecolor{tikzBoxColourNoResponse}{RGB}{117,147,176}
				\DeclareRobustCommand\tikzBoxNoResponse{\tikz\node[rectangle,fill=tikzBoxColourNoResponse,minimum width=\tikzBoxSize,minimum height = \tikzBoxSize,] (r) at (0,0) {};}
				
				\centering{}
				{\includegraphics[width=1.00\textwidth,keepaspectratio] {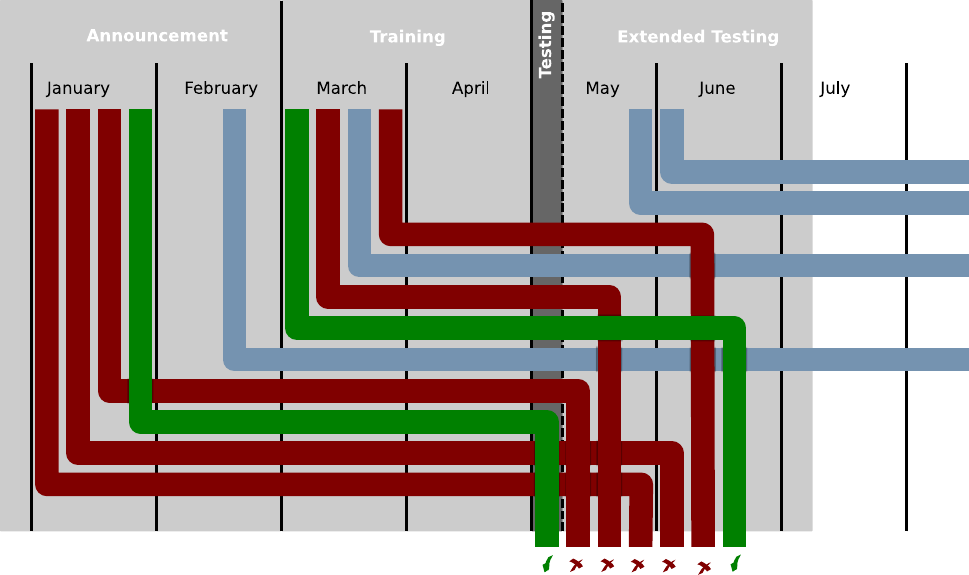}}
				\caption{\label{fig:parcitipants}Flow graph of participant registration and participation over the course of the challenge. From left to right time. Each colour coded line represents one participant: submitted green~(\tikzBoxSubmitted)\,\,ending in checkmark; withdrew red~(\tikzBoxWithdrew)\,\,ending in cross; no response blue~(\tikzBoxNoResponse)\,\,with horizontal line ending.}		
			\end{figure*}

			Figure \ref{fig:parcitipants} illustrates the registration and submission activity of the participants throughout the challenge period.
			
			After closing the \emph{training} phase, a total of eleven registrations were received. These registrations were distributed more or less evenly over the time span, with a slight increase following the challenge announcement and during and after the two attended conferences (see Section \ref{sec:challenge-organization}), where we explicitly promoted the participation of the challenge during the respective poster sessions. The interested parties and individuals had a global spread from Germany via Austria, Belgium, the United States, and Canada to Japan.
			
			Out of the eleven registrations, finally only one participant, referred to as \displayNameEngster, submitted a segmentation proposal within the testing phase. 			
			
			Another participant, referred to as \displayNameMichen, requested more time for the segmentation, which was granted due to the low submission count. Eventually, \displayNameMichen~submitted an automatic segmentation proposal eight weeks after the original challenge due date.
			
			We asked for comments from the remaining participants who did not submit a prediction, and received five responses. One participant stated that they had \enquote{underestimated the severity of the problem} and \enquote{that the task was not solvable with their existing resources and tools}. Two participants withdrew from the challenge due to \enquote{lack of time}, while a third participant changed employers during the course of the challenge and hence decided not to continue. Another participant stated he and his team withdrew from the challenge to invest company resources into another project. 
			
			Two additional participants registered after the end of the testing phase of the challenge. Due to the low participant count up to that point, their registrations were accepted, and they were granted immediate  access to the training and test data-sets. Each of these participants was given an extended time interval of four weeks to submit a segmentation proposal, but neither of them submitted. 						

		\subsection{Information on Selected Participating Teams}
			
			\subsubsection{Team One} \label{sec:results-engster}
\begin{figure*}
					\centering
					\subfloat[Slice wise instance segmentation using CDETR \label{fig:method-engster-pipe-2d}]				
					{\includegraphics[width=\textwidth,keepaspectratio]{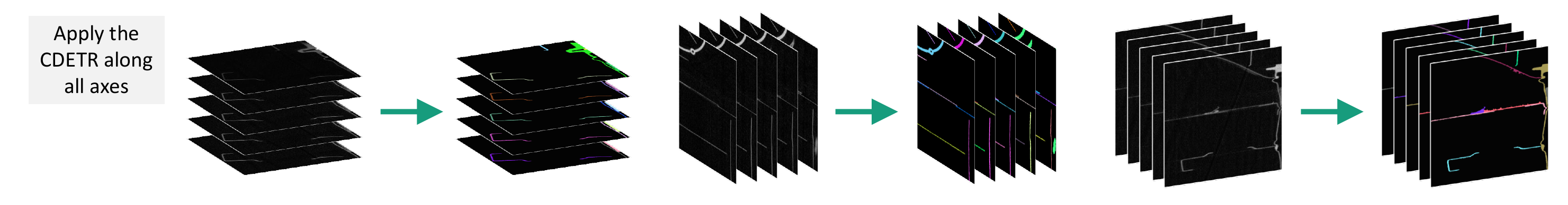}}
					\hfill
					\subfloat[Matching of line segments of intersecting orthogonal planes \label{fig:method-engster-pipe-match}]				
					{\includegraphics[width=\textwidth,keepaspectratio]{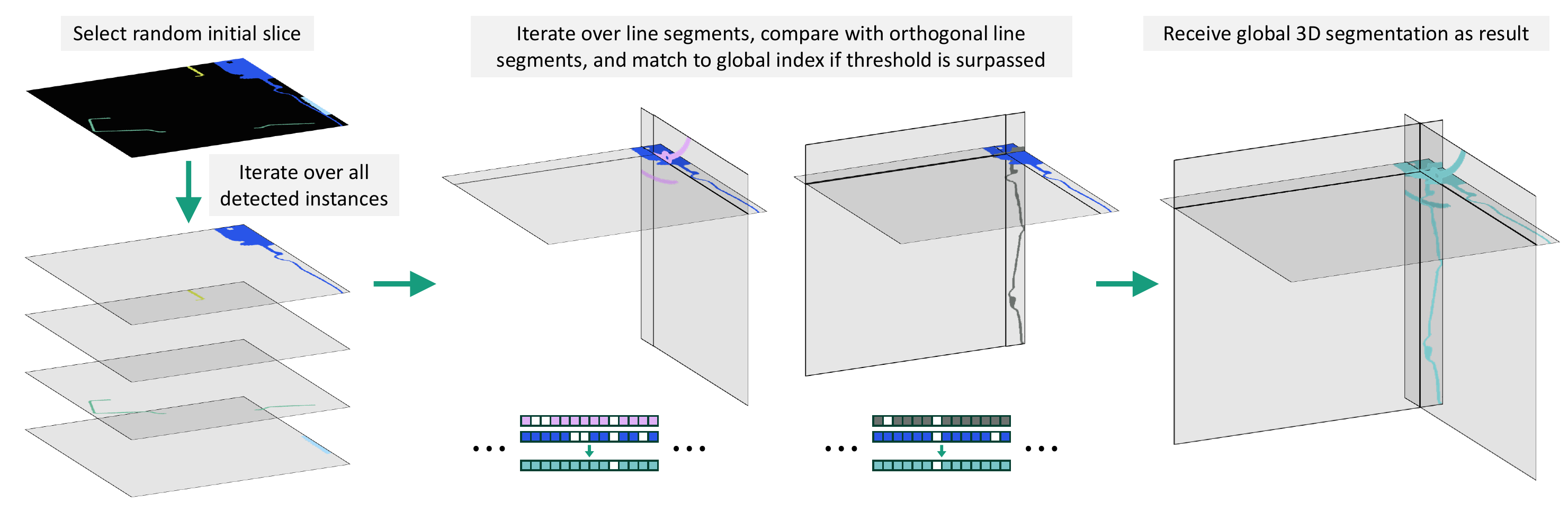}}
					\hfill
					\subfloat[Closing of line artefacts \label{fig:method-engster-pipe-postprocess}]				
					{\includegraphics[width=\textwidth,keepaspectratio]{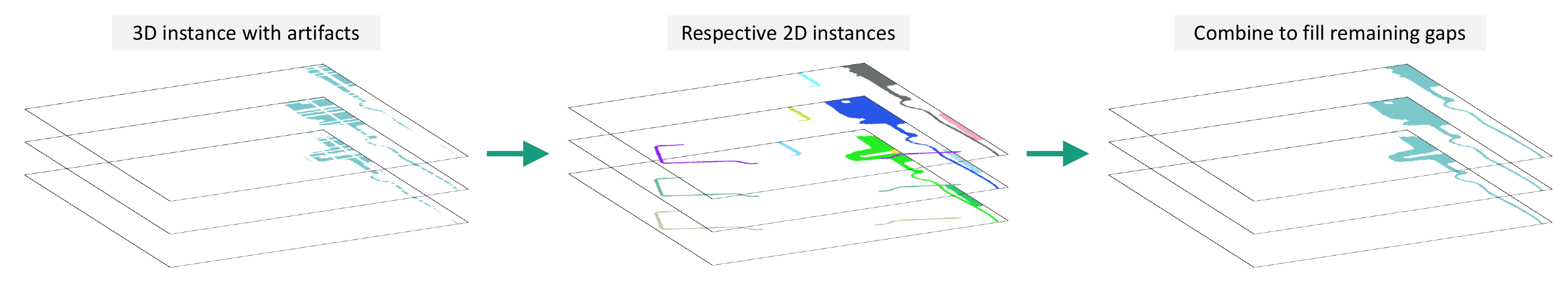}}	
					\caption{The proposed 3D instance segmentation pipeline by \displayNameEngster~which combines the conditional detection transformer (CDETR) for 2D instance segmentation with subsequent 3D-matching.}
					\label{fig:method-engster-pipe}
				\end{figure*}

				\displayNameEngster~was formed by JCE, NB, and MS. The contribution of \displayNameEngster~
				applied the conditional detection transformer (CDETR) \cite{cdetr} architecture. 
				The 2D instance segmentation masks obtained in a first step were combined into 3D instance segmentation through a novel matching algorithm in a second step.
				The complete pipeline is shown in Figure \ref{fig:method-engster-pipe}.
				
				\paragraph{Conditional Detection Transformers}
					The first step of \displayNameEngster\emph{'s} contribution utilized CDETR for 2D instance segmentation masks across all three planes (see Figure \ref{fig:method-engster-pipe-2d}).	The original Detection Transformer (DETR) model, originally proposed by Carion et al.~\cite{detr}, stands out as a notable architecture for object detection. DETR eliminates the need for hand-designed postprocessing techniques, such as non-maximum suppression, as its transformer inherently accounts for duplicate predictions. To obtain instance segmentation masks, a second head can be trained.	For \displayNameEngster\emph{'s} submission, a modified faster version, namely CDETR~\cite{cdetr}, which is related to conditional convolutional kernel generation, has been used. To increase the generalization performance, different data augmentations, including gaussian noise, colour jittering, and pixel dropout were utilized. Additionally, randomly stretched and skewed crops of the original images were applied.				
				
				\paragraph{3D-Matching}
					The next steps of the pipeline from \displayNameEngster~focused on 3D-matching of the individual 2D segmentations. Out of the available slices, an initial starting slice was selected. For each instance in this slice, the shared line segments with both other target planes were compared and matched (see Figure \ref{fig:method-engster-pipe-match}). If the degree of overlap between a line segment was above a defined threshold, the whole instance inside the target plane were set to a new global index. Since unsuccessful matches of segments resulted in line artefacts, the newly added global section were compared with the neighbouring slices of the start slice (see Figure \ref{fig:method-engster-pipe-postprocess}).  If an already added area overlapped with a found 2D instance in these slices, it was set to the same global index.				
				
				\paragraph{Post-Processing}
					Finally, several post-processing steps were applied to the obtained 3D instance segmentation, including morphological operations to close the remaining gaps.
					Furthermore, the initial 2D segments were reinserted into the matched 3D volume. For all 2D segments, the overlap with the 3D segments at the corresponding positions was calculated. When the overlap was above a predefined threshold, the regions were merged. That way the high accuracy of the 2D instance segmentation could be retained in the 3D volume.
		
				\paragraph{Conclusion}
					Team One contributed a novel approach that combined CDETR for 2D instance segmentation with a proposed 3D matching algorithm for 3D instance segmentation. The submitted resulting segmentation seems promising, particularly for larger components such as plates and valves. As an insufficient 2D instance segmentation can lead to a cumulative error and thus incorrect 3D matches, the model's generalization performance and inconsistencies in the reference annotations should be addressed. Furthermore, to reduce the computational complexity of the matching strategy, future work could focus on optimization and parallelization.	
			
			\subsubsection{Team Two}	\label{sec:results-michen}		
				\displayNameMichen\ (consiting of MM) employed a 3-class U-Net \cite{3D-U-Net, 7} model followed by postprocessing using classical image processing for the instance segmentation. The U-Net model had three classes: \emph{background}, \emph{object}, and \emph{border}. The \emph{border} class represented the regions between individual objects or between objects and the background. These \emph{border} regions were important for capturing the boundaries and transitions between the objects and the background, which were important for the instance segmentation. Additionally, postprocessing techniques were applied to improve the results. 
				
				The U-Net architecture was modified by replacing the encoder with an EfficientNet V2 model \cite{EfficientNetV2}. In addition, Mobile Inverted Bottleneck Convolution (MBConv) blocks \cite{MobileNetV2} were integrated into the decoder pathway of the U-Net. These MBConv blocks are responsible for feature upsampling and reconstruction. The inclusion of Squeeze-and-Excitation modules \cite{Squeeze} within these blocks helps to enhance the performance and accuracy of the U-Net model by recalibrating the channel-wise feature responses.
				
				\paragraph{Preprocessing}				
					Before feeding the data into the U-Net model, several pre-processing steps were performed. First the raw input data underwent total variation denoising \cite{Chambolle04}. Furthermore, the reference labels were transformed into the three classes (\emph{background}, \emph{object}, and \emph{border}) using morphological methods. However, it was later discovered that the choice of parameters resulted in thicker borders than desired, which was not satisfactory as multiple large metal sheets exhibit a thickness of only a few voxels.
				
				\paragraph{Training}				
					For training the 3-class U-Net model, about half of the available data-set volumes were used for training and the other half for validation. The model was trained using the AdamW \cite{68} optimizer with a learning rate of $3e-4$, weight decay of $2e-5$, and a dice cross-entropy loss function \cite{sudre2017generalised}. Class weighting was applied to emphasize the importance of the border class. Standard data augmentations such as contrast adjustment, noise addition, affine transformation, and introduction of artificial image artefacts were employed during training. The U-Net operated on blocks of $64^3$ with an overlap of 8\,voxels over the input volume.
				
				\paragraph{Post-processing}
					To convert the three-class output of the U-Net into the desired instance segmentation, a marker-based watershed algorithm was utilized alongside classical image processing techniques from the scikit-image library \cite{van2014scikit}. As mentioned previously, the U-Net prediction assigned a first class as foreground, a second class as background, and an additional third class for border regions. The watershed algorithm for image segmentation \cite{beucher1992watershed} was applied to fill and separate the object regions using the border class as a divider. The markers used for the watershed algorithm were obtained from the foreground class of the U-Net prediction. 
				
				\paragraph{Conclusion}
					The U-Net with watershed algorithm demonstrated good runtime performance, particularly for smaller parts, as it did not need expensive merging steps. However, it was observed that the algorithm erroneously merged larger metal sheets, possibly due to the thickness of the border pre-processing or excessive denoising. Further iterations of development and retraining, incorporating the suggested improvements, could hold the potential to enhance the overall performance. Exploring alternative methods for filling the border class, rather than relying on the watershed algorithm, may have the potential to yield improved results in computing an instance segmentation.	
					
		\subsection{Metric Values}		
	
\begin{figure}
				\centering
					\subfloat[\centering\displayNameEngster \label{fig:correlationMatrix-engster}]				
					{\includegraphics[width=0.5\columnwidth, interpolate=false,keepaspectratio]{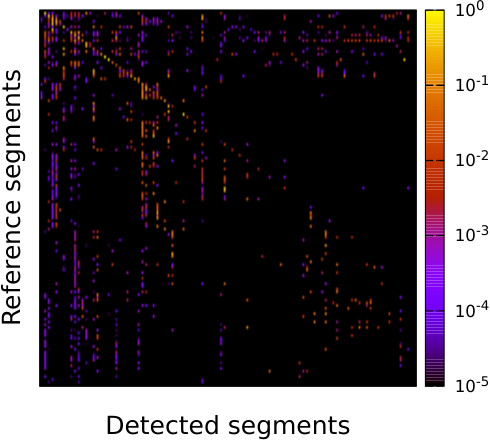}}	
					~\subfloat[\centering\displayNameMichen \label{fig:correlationMatrix-michen}]			
					{\includegraphics[width=0.5\columnwidth, interpolate=false,keepaspectratio]{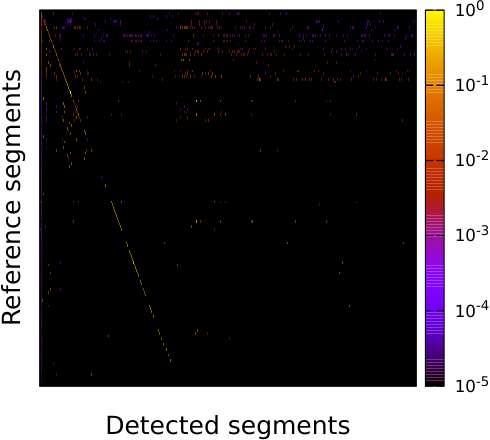}}

					\subfloat[\centering\displayNameEngster\, postprocessed and limited to main diagonal \label{fig:correlationMatrix-engster-connectedComponent-mainDiagonal}]
					{\includegraphics[width=0.5\columnwidth, interpolate=false,keepaspectratio]{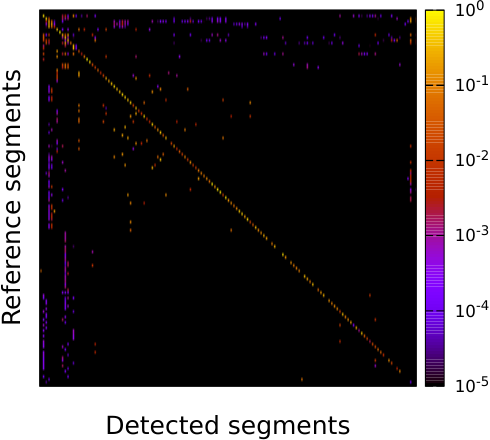}}
					~\subfloat[\centering\displayNameMichen\, postprocessed and limited to main diagonal \label{fig:correlationMatrix-michen-connectedComponent-mainDiagonal}]
					{\includegraphics[width=0.5\columnwidth, interpolate=false,keepaspectratio]{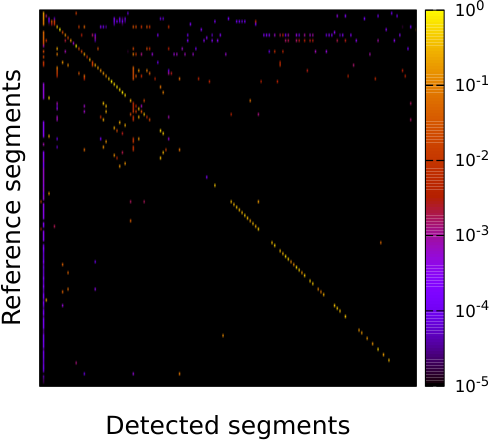}}		
				\caption{\label{fig:correlationMatrix}Correlation matrix of the segmentation of sub-volume $V_\text{test}$ (3072,~6656,~1024) between the reference annotation and the result from  \displayNameEngster\, Figure \ref{fig:correlationMatrix-engster} and \displayNameMichen\, Figure \ref{fig:correlationMatrix-michen}. Figures \ref{fig:correlationMatrix-engster-connectedComponent-mainDiagonal} and \ref{fig:correlationMatrix-michen-connectedComponent-mainDiagonal} show the correlation matrices after postprocessing and limited to the main diagonal for \displayNameEngster\,~and \displayNameMichen\,~respectively.}				
			\end{figure}

			To quantitatively evaluate the effectiveness of the instance segmentation approaches by \displayNameEngster\,~and \displayNameMichen\, an assessment of the correlation matrices (see Section \ref{methods-AssesmentMethod}) depicted in Figure \ref{fig:correlationMatrix} is performed to measure the performance of each method.
			
\begin{figure}
				\begin{centering}
					\subfloat[\label{result-failuremode-engster-a}]			
					{\includegraphics[width=0.45\columnwidth,keepaspectratio]{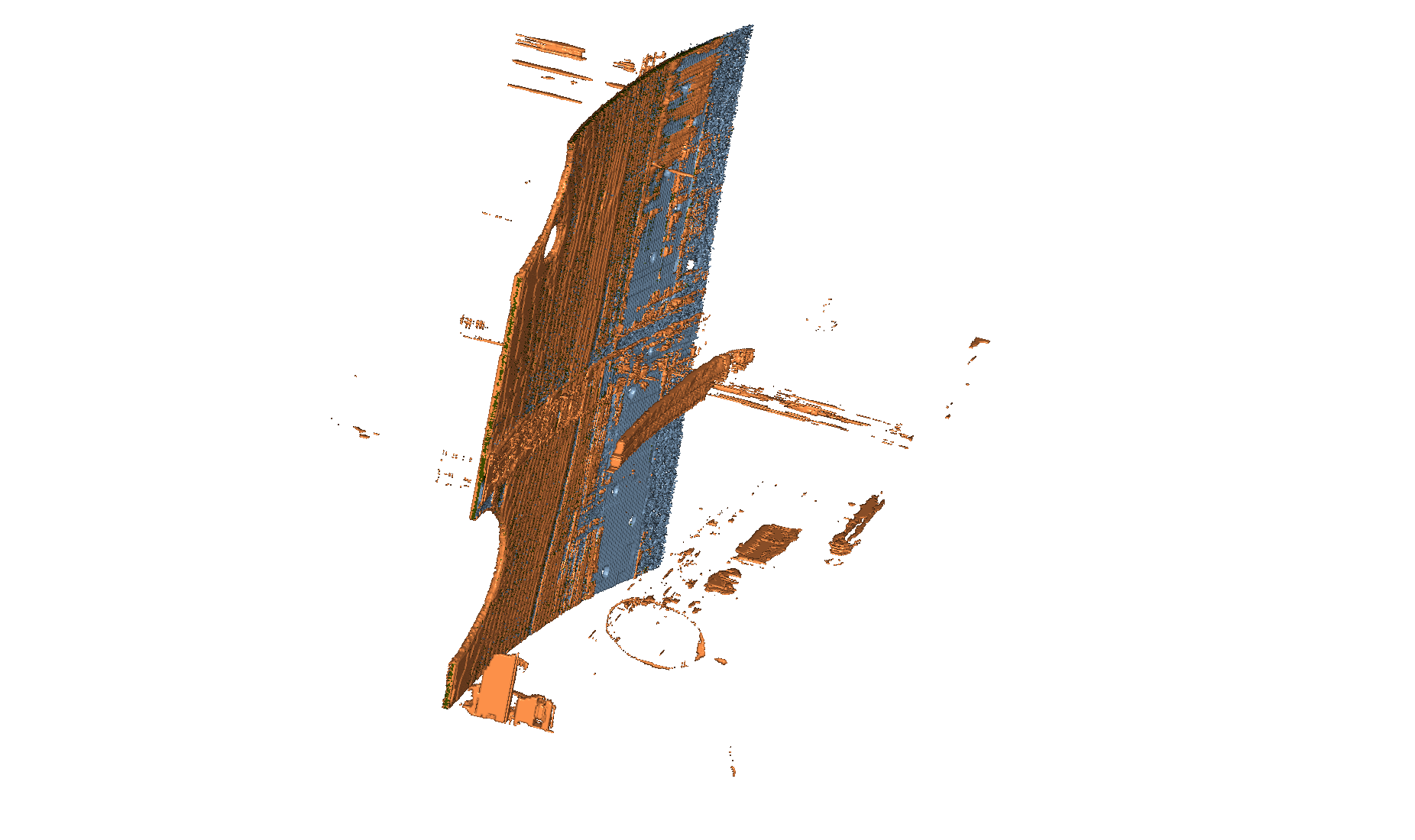}}	
					~\subfloat[\label{result-failuremode-engster-b}]
					{\includegraphics[width=0.45\columnwidth,keepaspectratio]{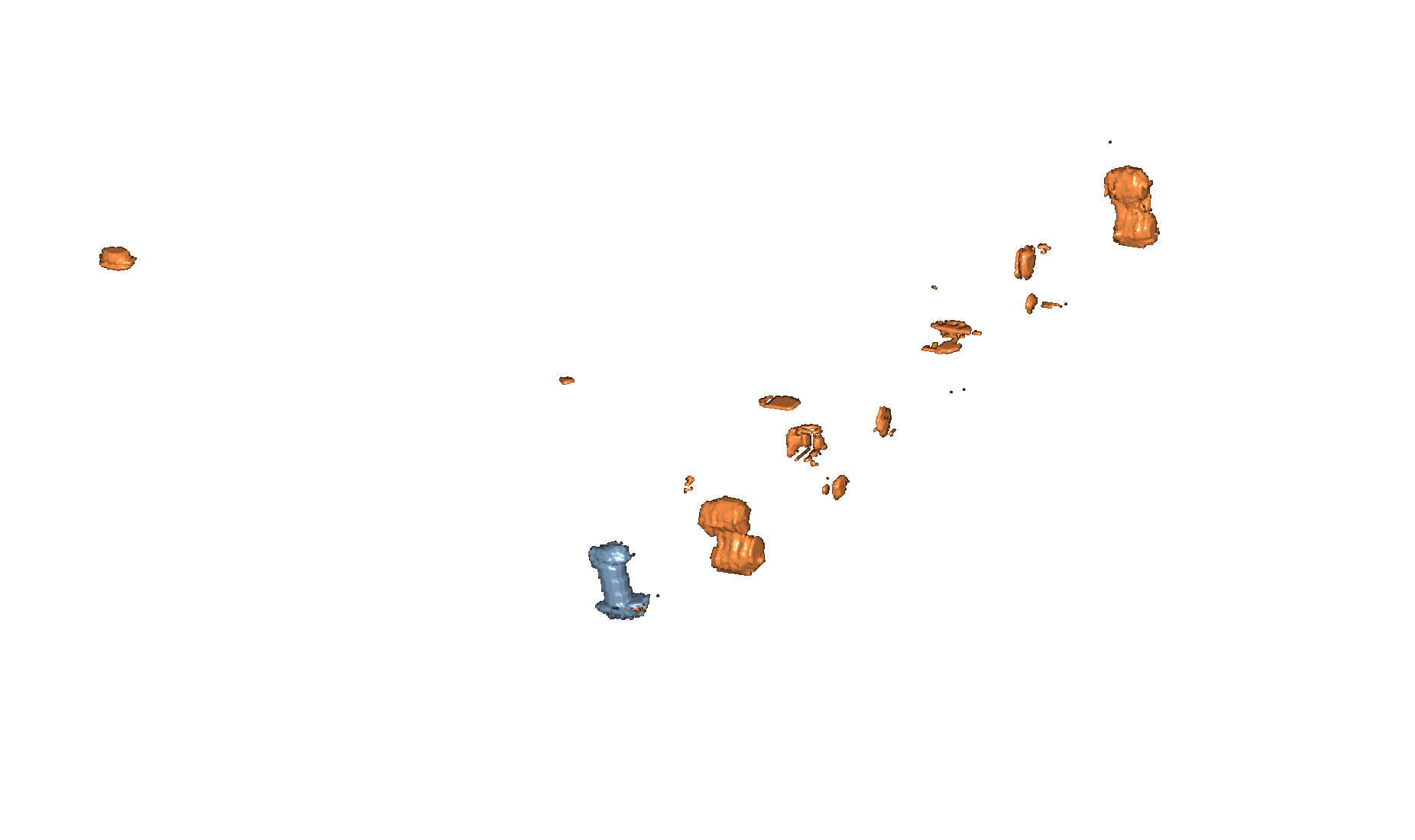}}	
				\end{centering}			
				\caption{\label{fig:result-failuremode-engster}Renderings of typical error cases for the \displayNameEngster\, algorithm, where segments are falsely joined or segmented together. For example, in Figure \ref{result-failuremode-engster-a}, the algorithm tends to spread a segment over multiple non-connected segments, such as orthogonal and parallel metal sheets. Additionally, in Figure \ref{result-failuremode-engster-b}, the algorithm may incorrectly segment multiple rivets together. To account for these error cases and establish a better comparison between detected and reference segments, a connected component analysis was applied. Colour coded: reference segments blue~(\tikzBoxFalseNegative); proposed segmentation of \displayNameEngster\,orange~(\tikzBoxFalsePositive); overlap green~(\tikzBoxTruePositive), not visible in these renderings as it corresponds to the centre of the segmentations.}				
			\end{figure}

			The \displayNameEngster\, method (see Figure \ref{fig:correlationMatrix-engster}) performed well for reference segments with large voxel counts in the upper-left corner of the matrix but failed to segment smaller reference segments (as indicated by the discontinuity of the diagonal line in the middle and right sections of the matrix). A typical error case is shown in Figure \ref{fig:result-failuremode-engster}, where the \displayNameEngster\, method combined multiple segments. Given that the correlation matrix was significantly affected by this failure mode, we conducted a postprocessing step involving connected component analysis of the segments. This also helped to reduce the under-segmentation of the reference segments, which is primarily visible as vertical lines in the left portion of Figure \ref{fig:result-failuremode-engster}. This postprocessing step increased the initial count of proposed segments from 103 to 957, as multiple detected segments were incorrectly combined. Therefore, Figure \ref{fig:correlationMatrix-engster-connectedComponent-mainDiagonal} depicts the cropped correlation matrix of \displayNameEngster\, after the connected component analysis. The region surrounding the main diagonal is of particular interest, as it displays proposed segments in conjunction with their corresponding reference segments.
			
			In comparison, Figure \ref{fig:correlationMatrix-michen} illustrates the complete correlation matrix containing 384 segments from the segmentation proposal by \displayNameMichen, whereas Figure \ref{fig:correlationMatrix-michen-connectedComponent-mainDiagonal} displays a refined version confined to the main diagonal. We also conducted the aforementioned  postprocessing step on this segmentation proposal to ensure fair comparability between the algorithms. However, the results of the \displayNameMichen\, algorithm remained mostly unaffected, with an initial count of 371 proposed segments.	Notably, the \displayNameMichen\, proposal typically does not exhibit the same failure mode as the algorithm from \displayNameEngster\, and demonstrated a lesser tendency to over-segment the large reference objects which can be seen by the relative sparse nature of the correlation matrix. However, a common failure mode was under-segmentation, as the algorithm from \displayNameMichen\, combined most of the large metal sheets into one single segment. Nevertheless, the algorithm exhibited its strength in segmenting smaller objects such as screws, which is reflected in the high values along the lower section of the main diagonal of the correlation matrix, albeit with discontinuities. The algorithm also exhibits a low count of outliers in the lower half of the correlation matrix.
			
			\begin{figure*}
				\centering{}
				{\includegraphics[width=1.00\textwidth,keepaspectratio] {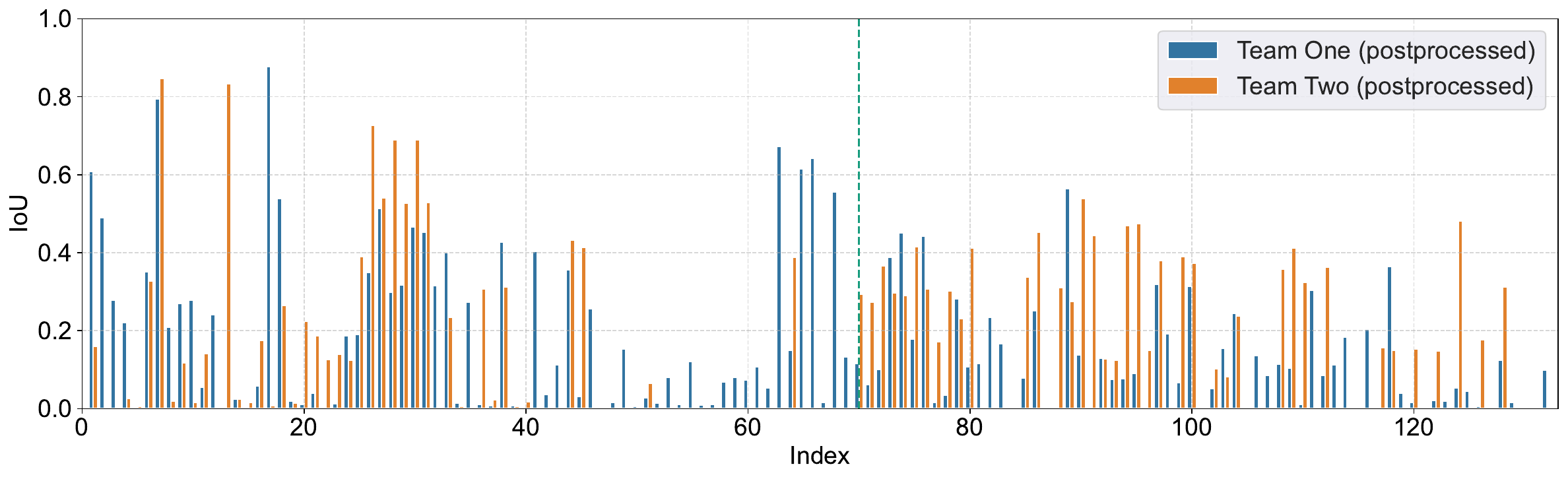}}
				\caption{\label{fig:correlationMatrix-mainDiagonal-plot} Bar plot of the main diagonals from the correlation matrix of \displayNameEngster~and \displayNameMichen~(Figures \ref{fig:correlationMatrix-michen-connectedComponent-mainDiagonal} and  \ref{fig:correlationMatrix-michen-connectedComponent-mainDiagonal}). The green vertical line indicates the division between the front and back half of the main diagonal, which contain large and small segments respectively.}		
			\end{figure*}	
			
			Figure \ref{fig:correlationMatrix-mainDiagonal-plot} shows a plot of the correlation matrix main diagonal for the postprocessed segmentation proposal by \displayNameEngster\, and \displayNameMichen sorted by reference voxel count from high to low. It can be observed that \displayNameEngster\, exhibits good segmentation quality, particularly in the left region where segments with a large count of voxels are located. Both algorithms demonstrate enhanced results for different segments. Towards the segments with lower voxel count, the \displayNameMichen\, algorithm tends to demonstrate slightly better segmentation quality than the \displayNameEngster\, algorithm. 
			
			Compared to the reference segmentation of the testing sub-volume $V_\text{test}$, the submitted results achieved quite good results in terms of over- and under-segmentation. Additionally, the results of the proposed segmentation methods are comparable to our own prior approach \cite{Gruber2020}, which was not included in the challenge, due to having an extended background knowledge about the scanning process as well as the annotation methods used, which would have resulted in an unfairness to the challenge participants.	
			
		\subsection{Rankings}
			
			Table \ref{tab:correlationMatrix-mainDiagonal-statistics} provides a quantitative evaluation of the main diagonals from \displayNameEngster\, and \displayNameMichen\, in terms of the proposed segmentations. The evaluation is categorized based on the front or back half of the main diagonal and highlights their remarkably close proximity to each other. Regarding all segments, both \displayNameEngster\, and \displayNameMichen\, achieved the comparable mean IoU scores. For large segments, \displayNameEngster\, (after postprocessing) achieved the highest mean IoU with 0.20. Concerning small segments, \displayNameMichen\, achieved the highest mean IoU with 0.16, independent of postprocessing.	
			Hence, ranking the results is challenging as each algorithm has its own advantages and shortcomings, making a unified ranking difficult. For real-world applications, as well as within this challenge, none of the proposed two algorithms can be deemed superior over the other. Also, the purpose of this challenge was not primarily focused on ranking algorithms, but rather to foster collaboration between scientists from different fields and explore new approaches for the NDT field.
			
			\begin{table*}
				\centering
				\begin{adjustbox}{max width=\textwidth}
					\begin{tabular}{rlcccc} \toprule
						\multicolumn{2}{c}{\makecell{Correlation Matrix}} 		& \displayNameEngster 	& \makecell{\displayNameEngster\\ (postproceced)} 	& \displayNameMichen & \makecell{\displayNameMichen\\ (postproceced)}	\\ \midrule				
						\multirow{3}{*}{All Segments} 		& Maximum			& 0.87			 		& \bf{0.88}											& 0.85				 & 0.85			\\ 
															& Mean				& 0.04			 		& \bf{0.16}											& 0.15			 	 & 0.15			\\ 
															& Standard Deviation& \bf{0.12}		 		& 0.19												& 0.20				 & 0.20     	\\ \midrule
						\multirow{3}{*}{Large Segments} 	& Maximum			& 0.87			 		& \bf{0.88}											& 0.85				 & 0.85			\\ 
															& Mean				& 0.08			 		& \bf{0.20}											& 0.19				 & 0.14			\\ 
															& Standard Deviation& \bf{0.17}	 			& 0.22												& 0.23				 & 0.23    		\\ \midrule
						\multirow{3}{*}{Small Segments} 	& Maximum	 		& 0.003		 	 		& \bf{0.56}											& 0.54				 & 0.54			\\ 
															& Mean				& 0.000			 		& 0.11												& \bf{0.16}			 &\bf{0.16}		\\ 
															& Standard Deviation& \bf{0.000}	 	 	& 0.13												& 0.17				 & 0.17     	\\ \bottomrule
					\end{tabular}	
				\end{adjustbox}
					
				\caption{\label{tab:correlationMatrix-mainDiagonal-statistics} Key metrics of the main diagonal of the correlation matrices in Figure\,\ref{fig:correlationMatrix} for the test sub-volume $V_\text{test}$.}
			\end{table*}
				
		\subsection{Further Analyses}
			
			This section focuses on a more detailed qualitative analysis of the proposed segmentations, with the main objective of highlighting their unique and common advantages, biases, and shortcomings, which cannot be gleaned from the correlation matrix in Figure \ref{fig:correlationMatrix} alone.
			
\begin{figure*}
				\begin{centering}
					\subfloat[Reference \label{fig:result-all-reference}]					
					{\includegraphics[width=0.33\textwidth,keepaspectratio]{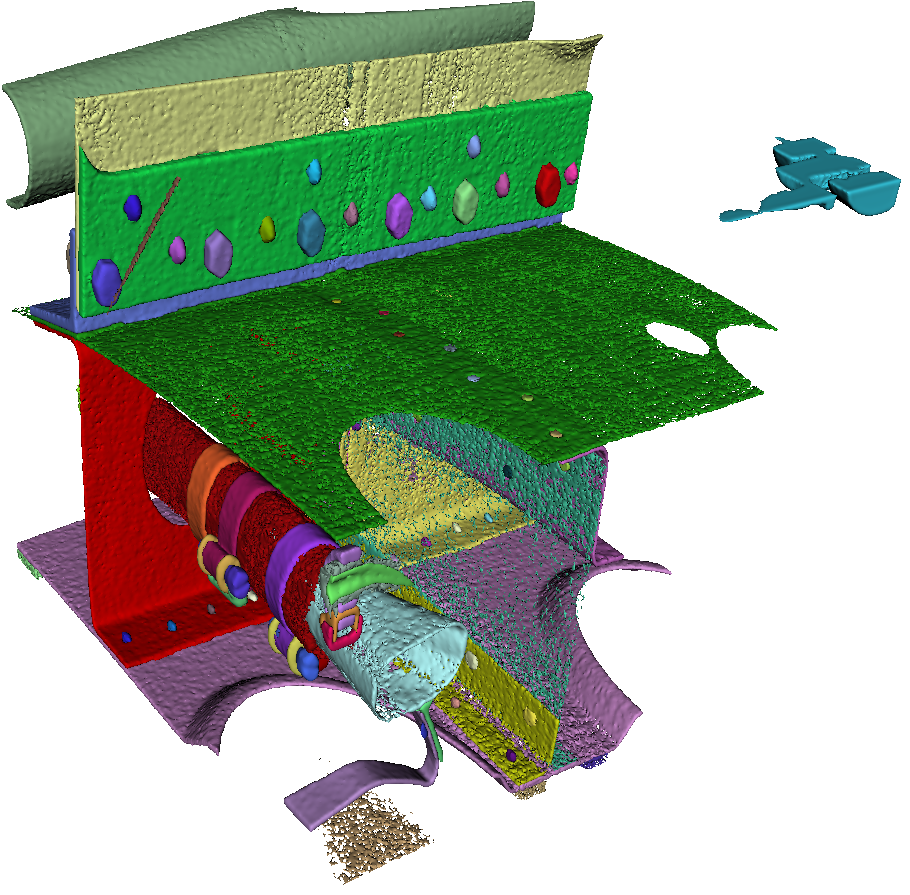}}	
					~\subfloat[\displayNameEngster \label{fig:result-all-engster}]
					{\includegraphics[width=0.33\textwidth,keepaspectratio]{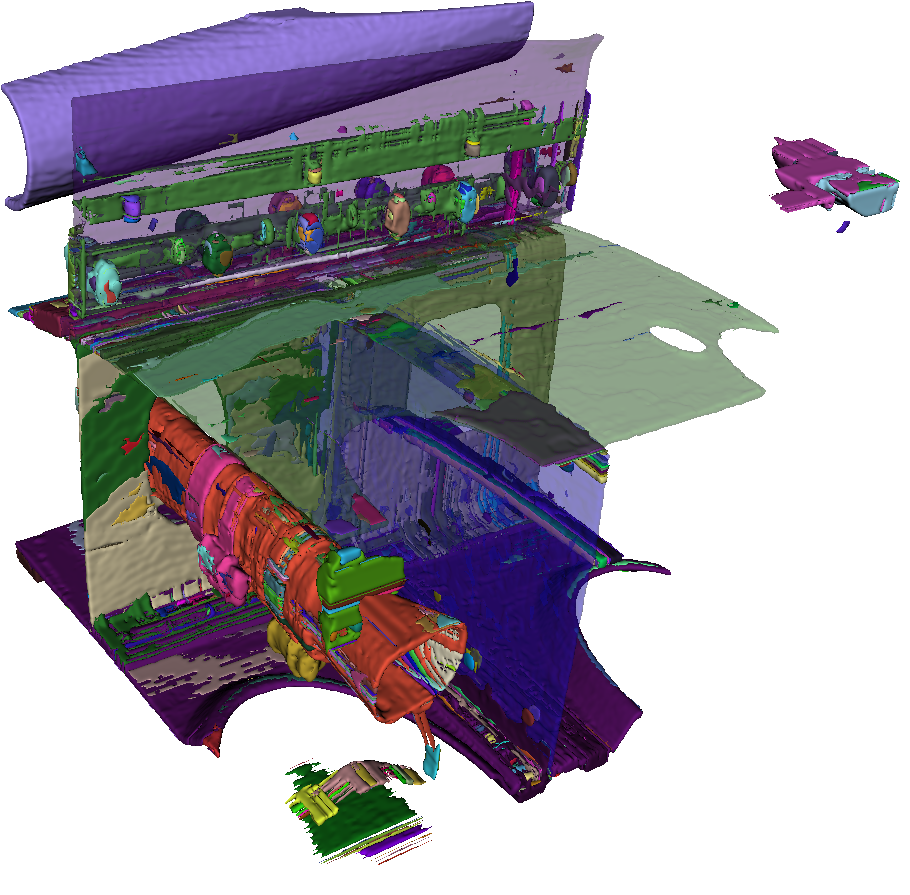}}	
					~\subfloat[\displayNameMichen \label{fig:result-all-michen}]
					{\includegraphics[width=0.33\textwidth,keepaspectratio]{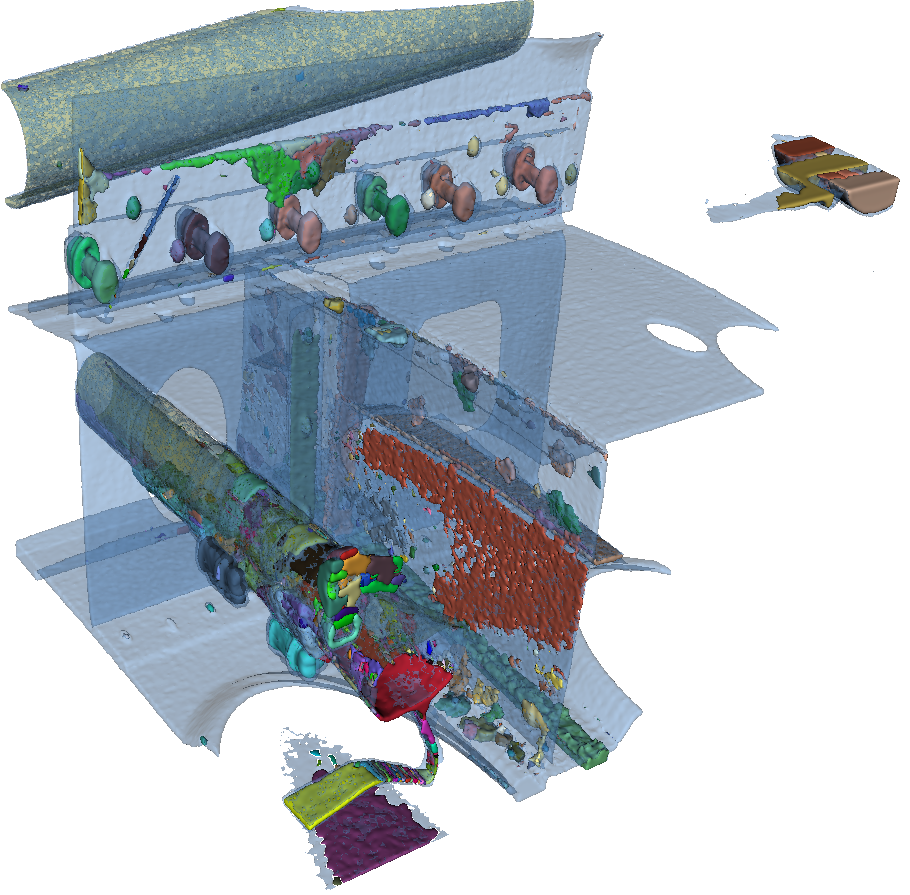}}
				\end{centering}
				\caption{\label{fig:result-all} Renderings of the reference (Figure \ref{fig:result-all-reference}) and proposed segmentations by \displayNameEngster~(Figure \ref{fig:result-all-engster}) and \displayNameMichen~(Figure \ref{fig:result-all-michen}) after connected component analysis.}
			\end{figure*}

			Figure \ref{fig:result-all} shows 3D renderings of the reference segmentation and the complete proposed segmentation by \displayNameEngster\, and \displayNameMichen\, after connected component analysis. Given the large count of segments involved, it is important to acknowledge the limitations of this display method. Specifically, the use of colour coding for the segments may result in misunderstandings due to similar colours or overlapping segments. Furthermore, even if the interior of segments has been correctly segmented, this core can be obscured by voxels located further outside, rendering it invisible.
			
\begin{figure}
		 		\centering
		 		\subfloat[IoU~0.61]{ \label{fig:result-big-engster-connectedComponent-reference31-proposed306}\includegraphics[width=0.25\columnwidth,keepaspectratio]{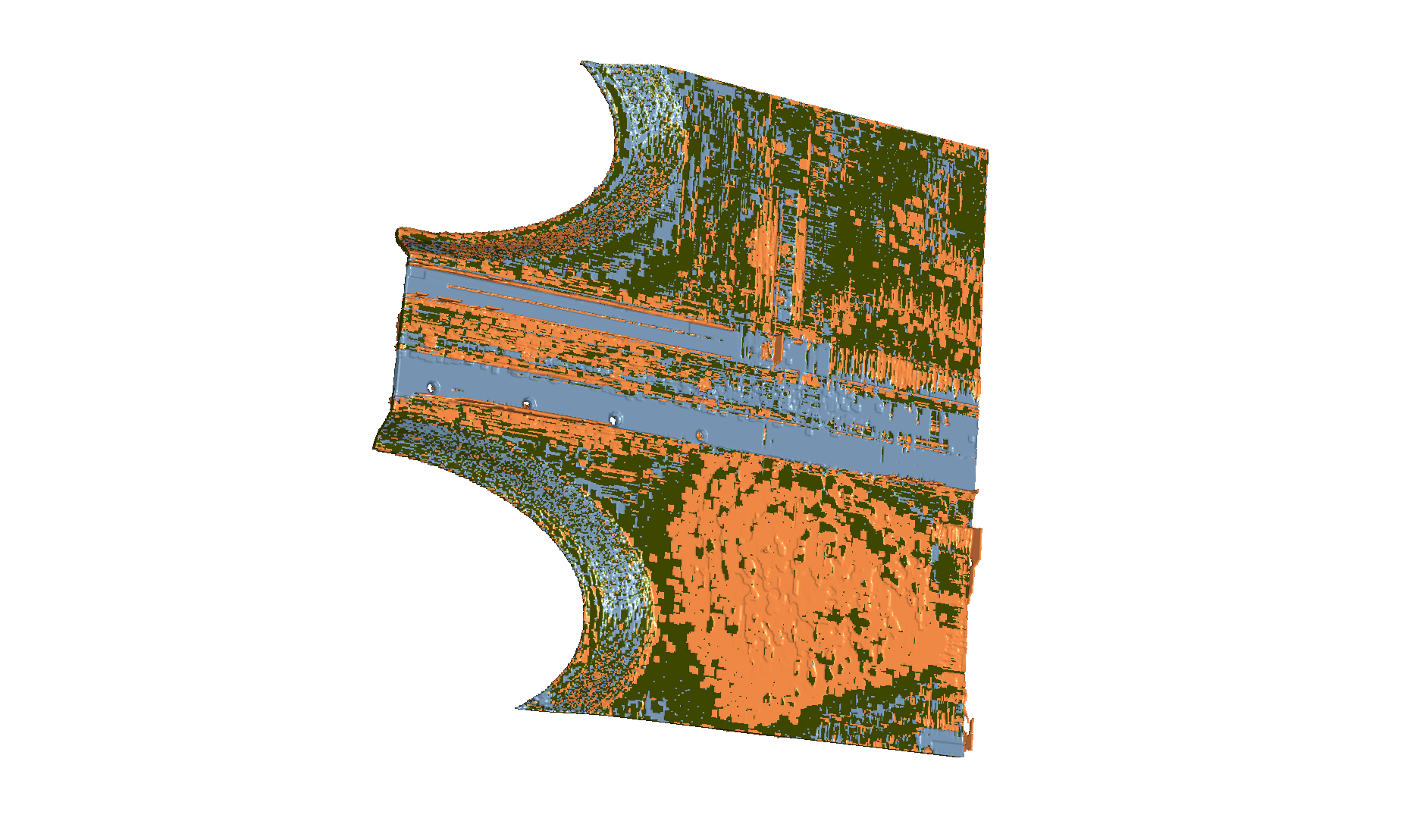}}
		 		~\subfloat[IoU~0.16 ]{\label{fig:result-big-michen-connectedComponent-reference31-proposed1}\includegraphics[width=0.25\columnwidth,keepaspectratio]{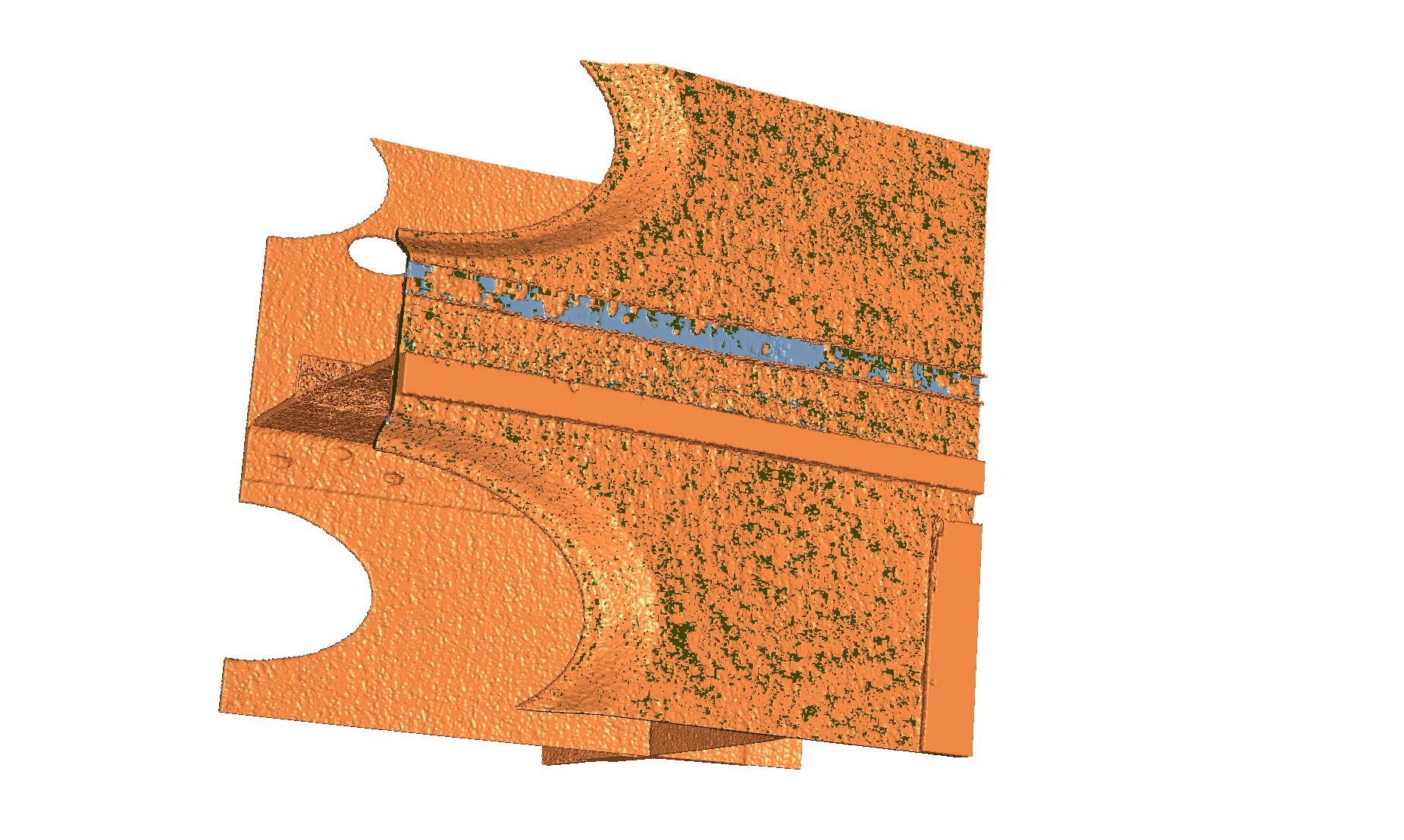}}
		 		~\subfloat[IoU~0.49 ]{\label{fig:result-big-engster-connectedComponent-reference30-proposed597}\includegraphics[width=0.25\columnwidth,keepaspectratio]{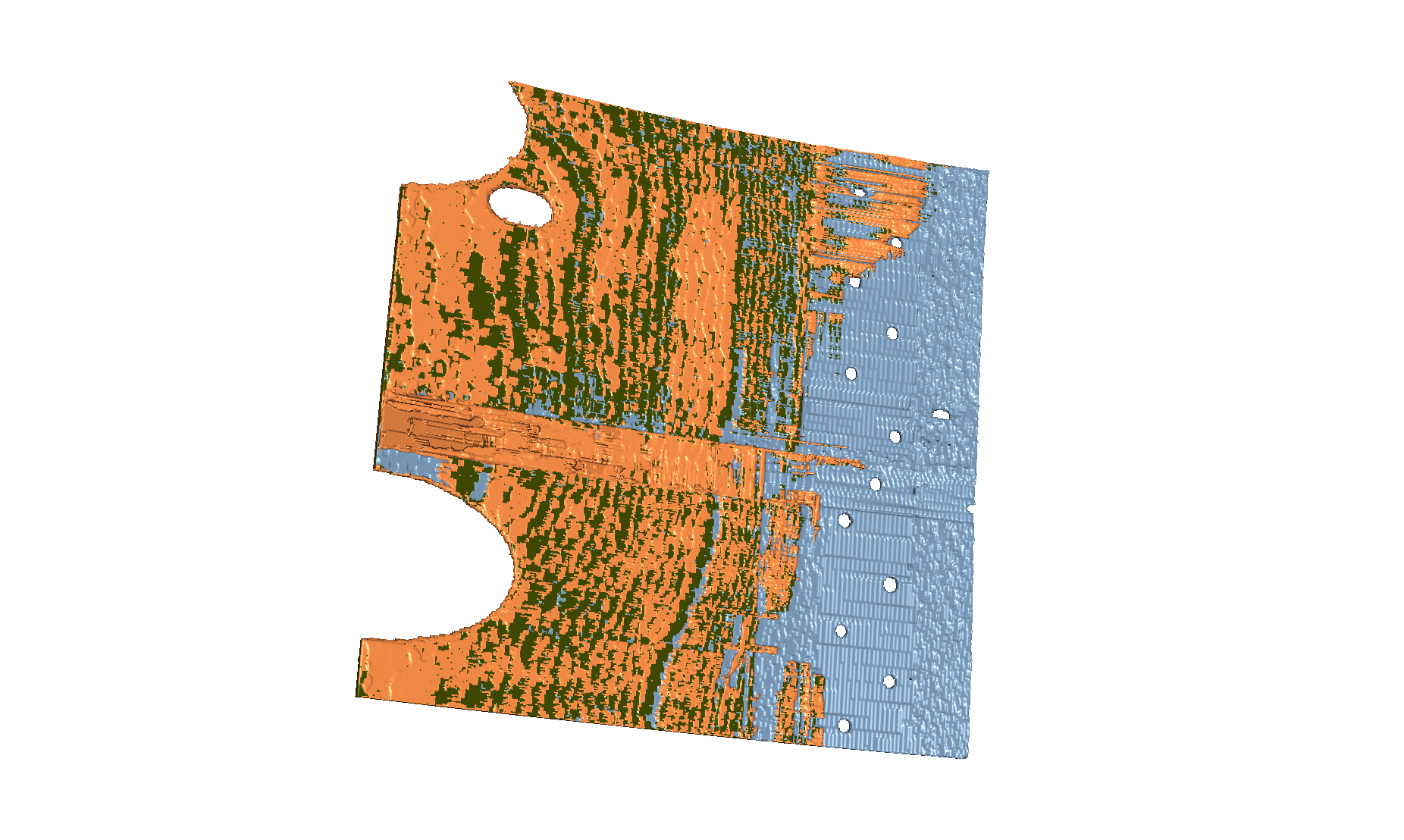}}
		 		~\subfloat[IoU~0.002]{\label{fig:result-big-michen-connectedComponent-reference30-proposed214}\includegraphics[width=0.25\columnwidth,keepaspectratio]{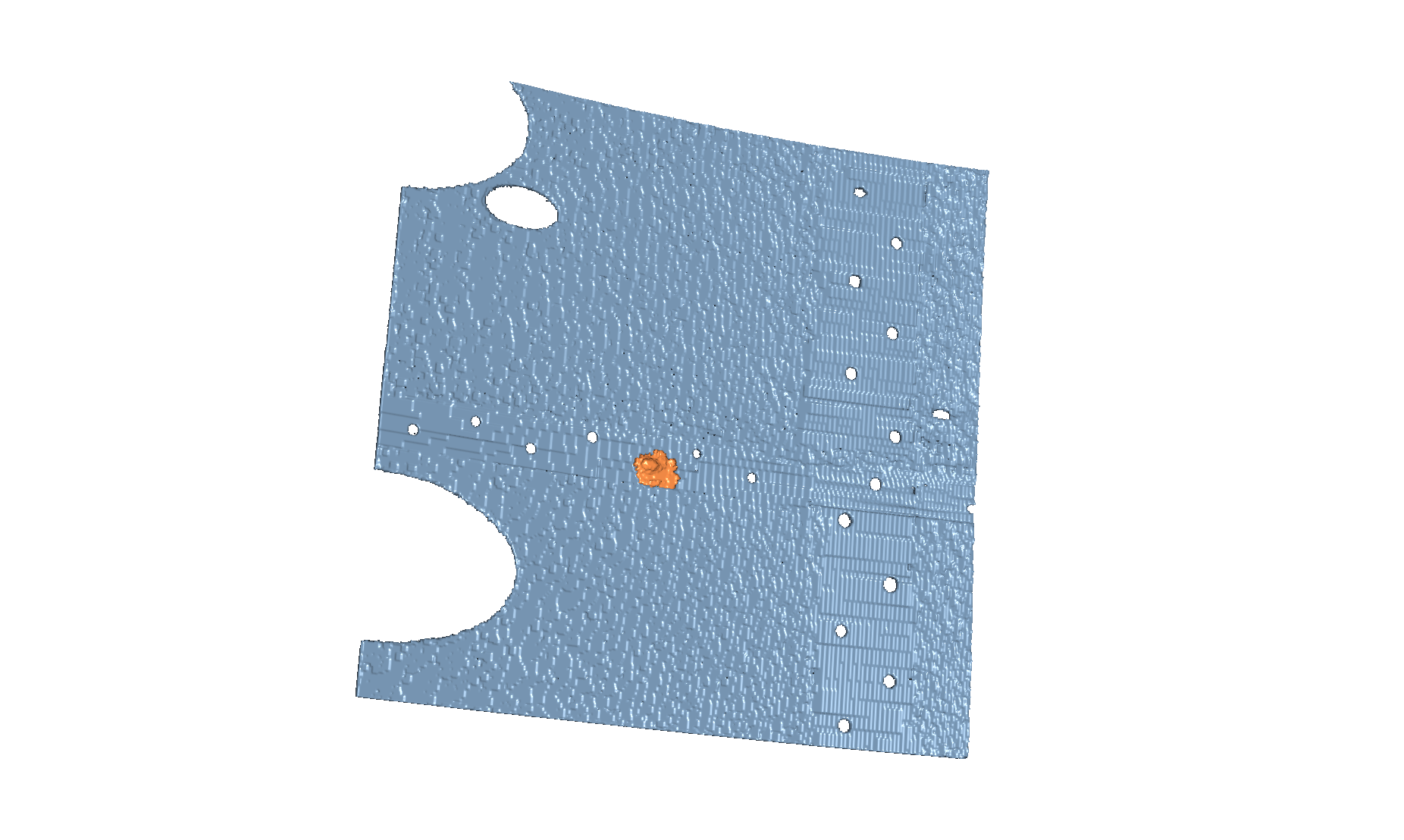}}
		 		
		 		\subfloat[IoU~0.28]{ \label{fig:result-big-engster-connectedComponent-reference24-proposed450}\includegraphics[width=0.25\columnwidth,keepaspectratio]{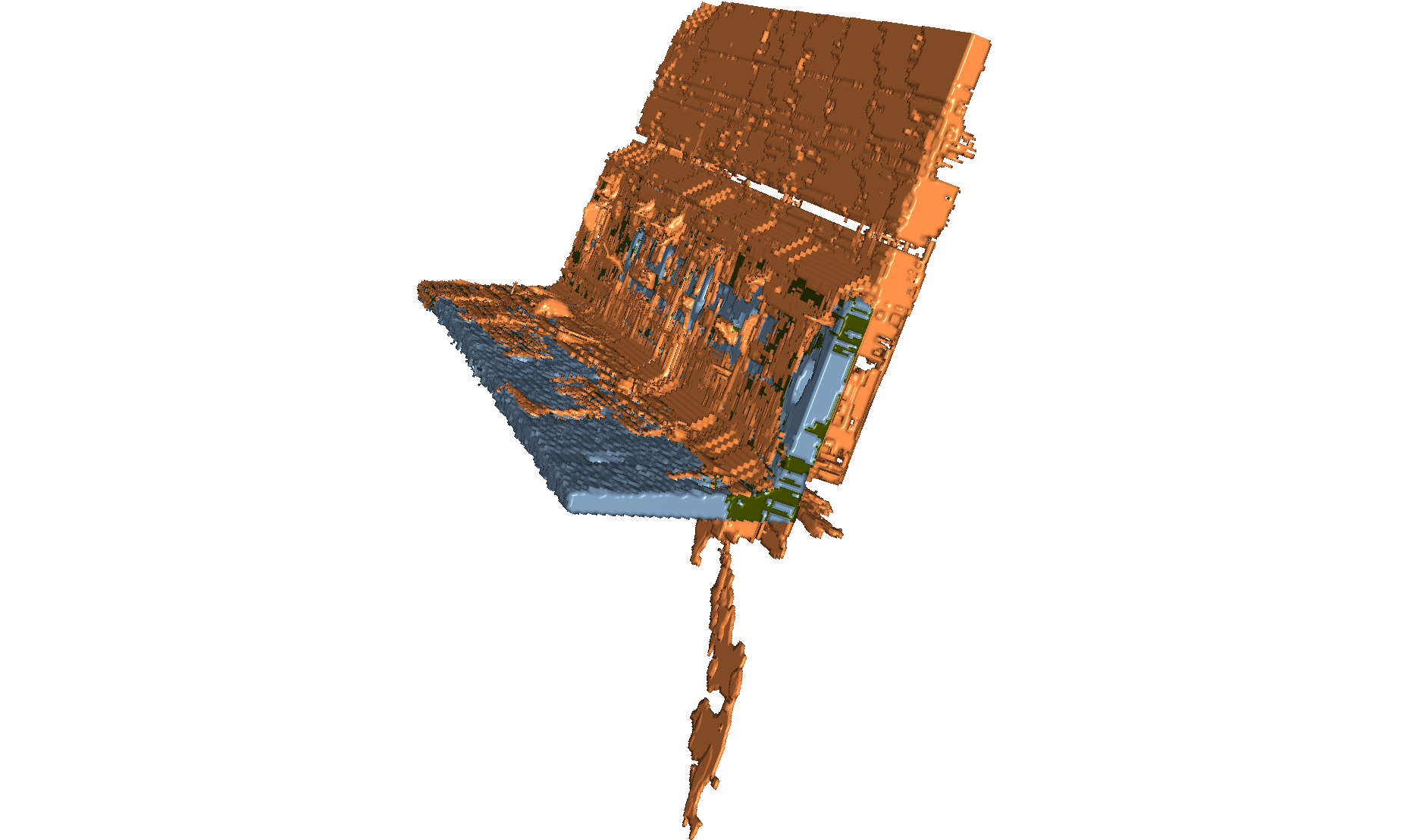}}
		 		~\subfloat[IoU~0.0003 ]{\label{fig:result-big-michen-connectedComponent-reference24-proposed208}\includegraphics[width=0.25\columnwidth,keepaspectratio]{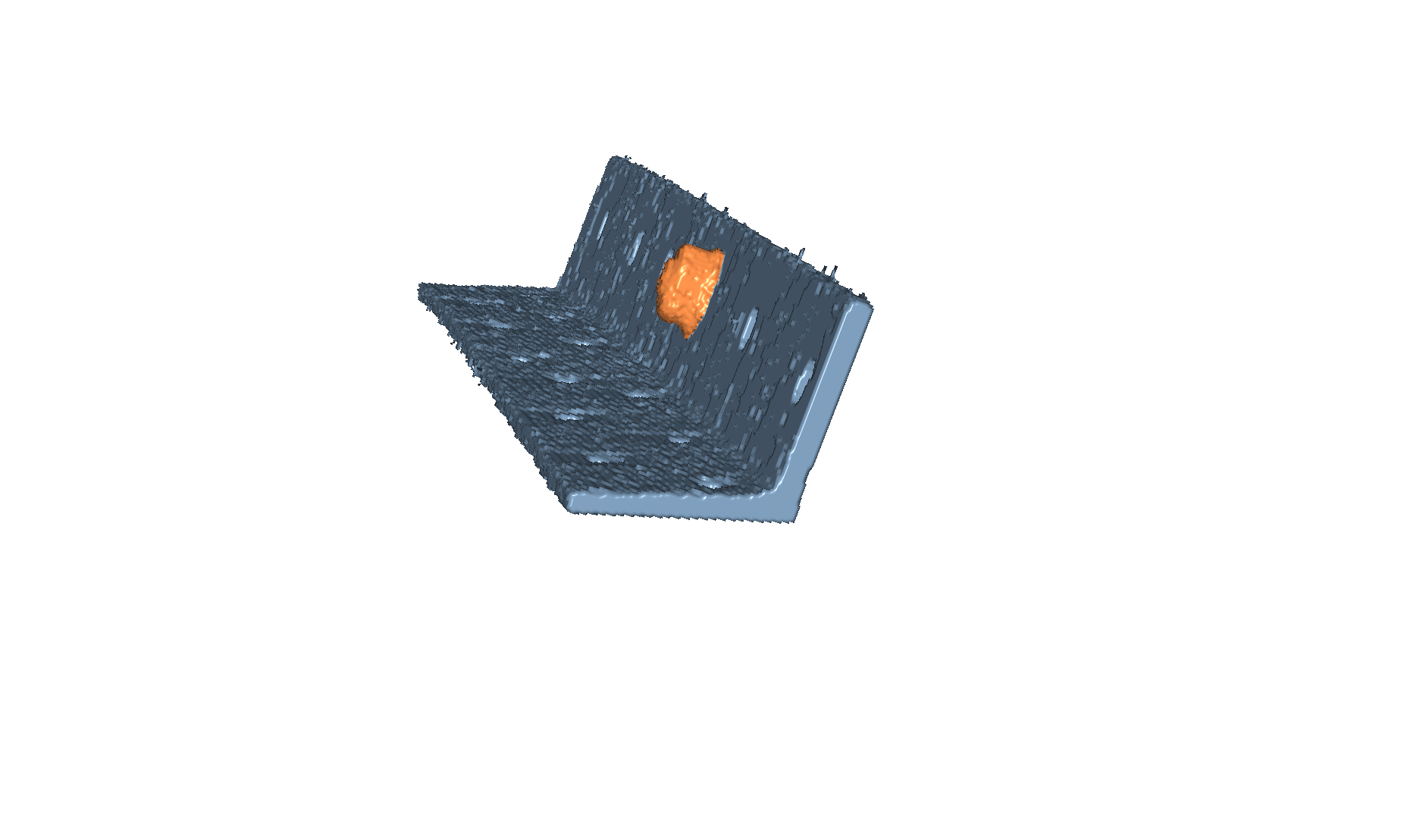}}
		 		~\subfloat[IoU~0.22 ]{\label{fig:result-big-engster-connectedComponent-reference23-proposed363}\includegraphics[width=0.25\columnwidth,keepaspectratio]{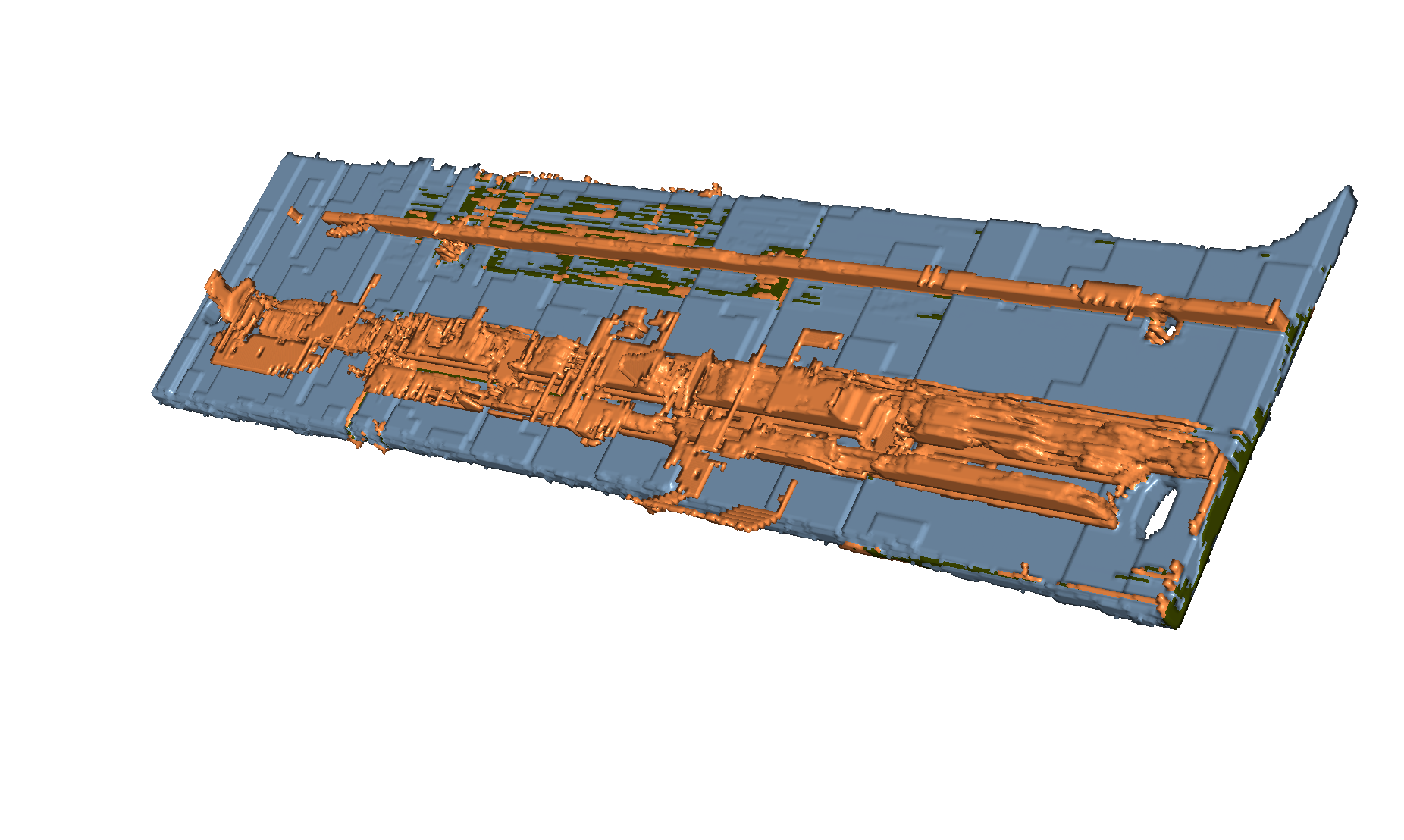}}
		 		~\subfloat[IoU~0.02]{\label{fig:result-big-michen-connectedComponent-reference23-proposed99}\includegraphics[width=0.25\columnwidth,keepaspectratio]{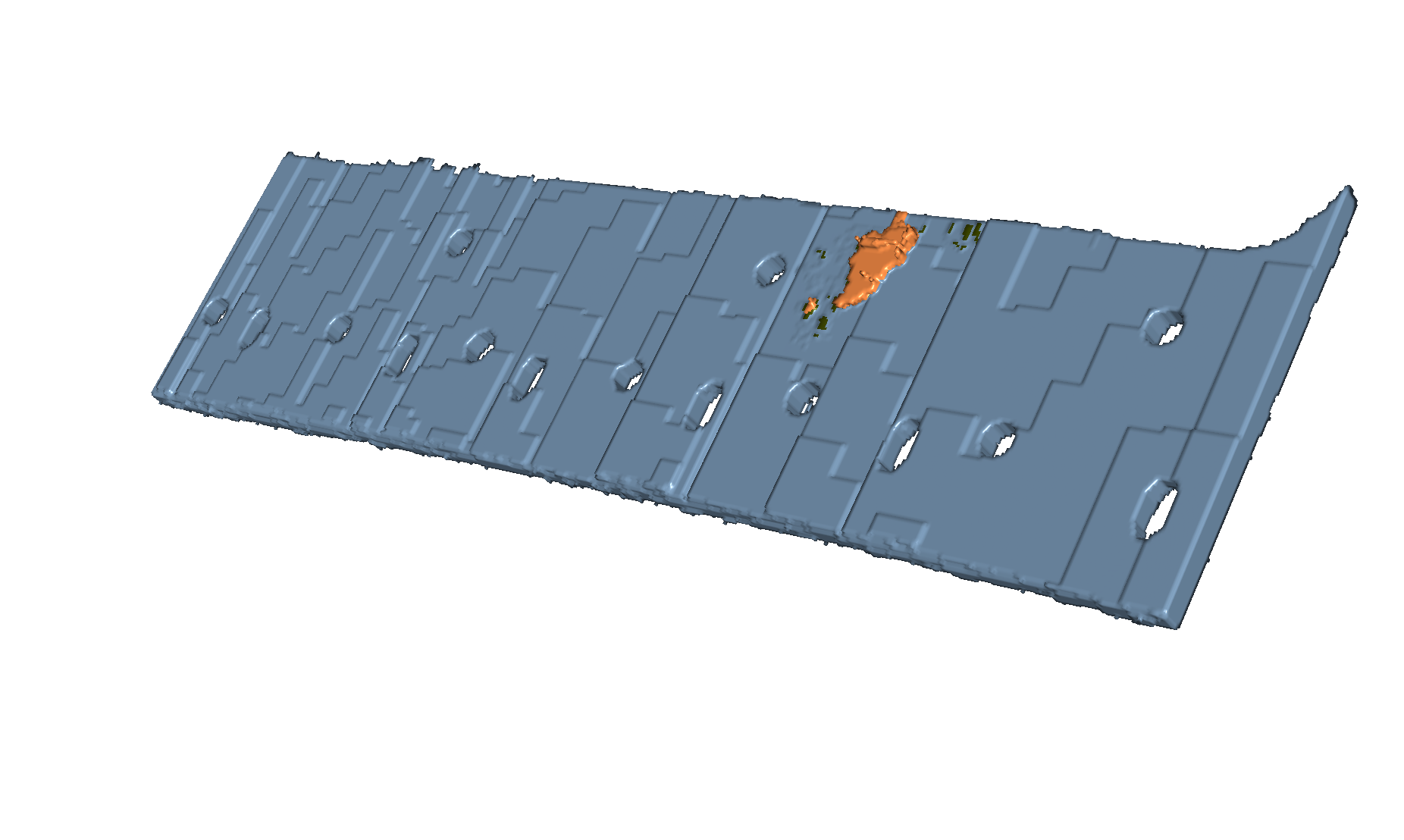}}	
		 		
		 		\subfloat[IoU~0.003]{ \label{fig:result-big-engster-connectedComponent-reference25-proposed417}\includegraphics[width=0.25\columnwidth,keepaspectratio]{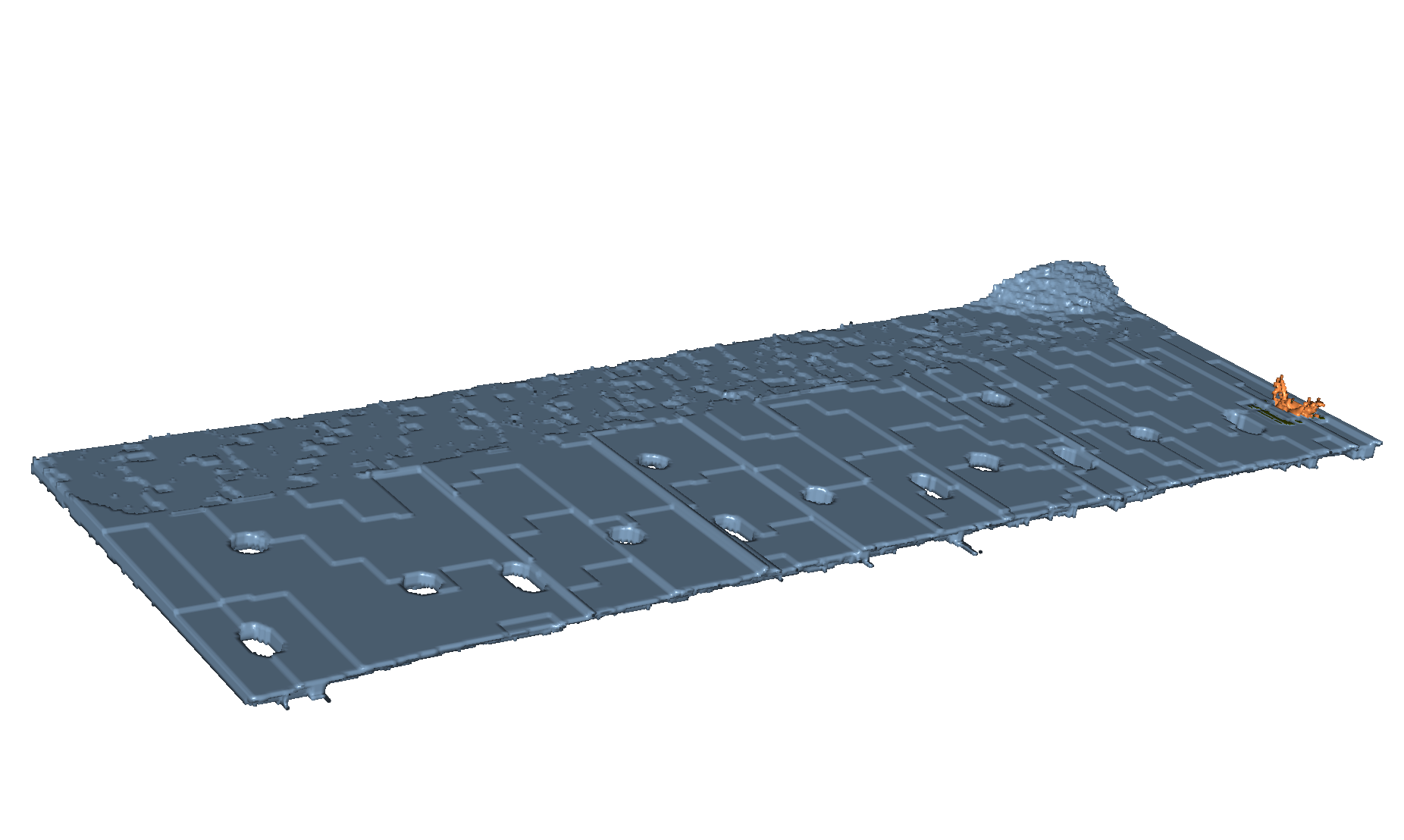}}
		 		~\subfloat[IoU~0.005 ]{\label{fig:result-big-michen-connectedComponent-reference25-proposed225}\includegraphics[width=0.25\columnwidth,keepaspectratio]{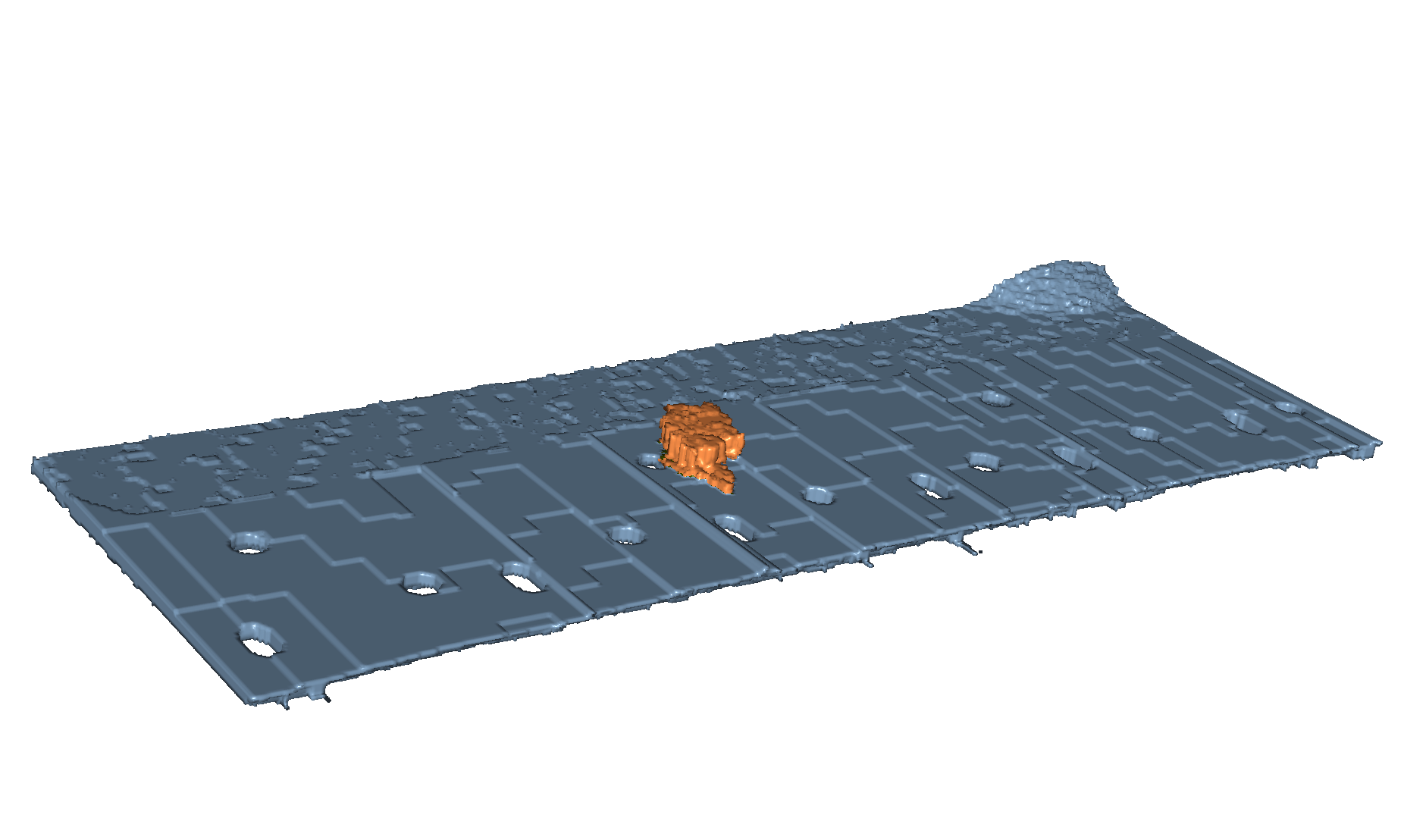}}
		 		~\subfloat[IoU~0.35 ]{\label{fig:result-big-engster-connectedComponent-reference124-proposed883}\includegraphics[width=0.25\columnwidth,keepaspectratio]{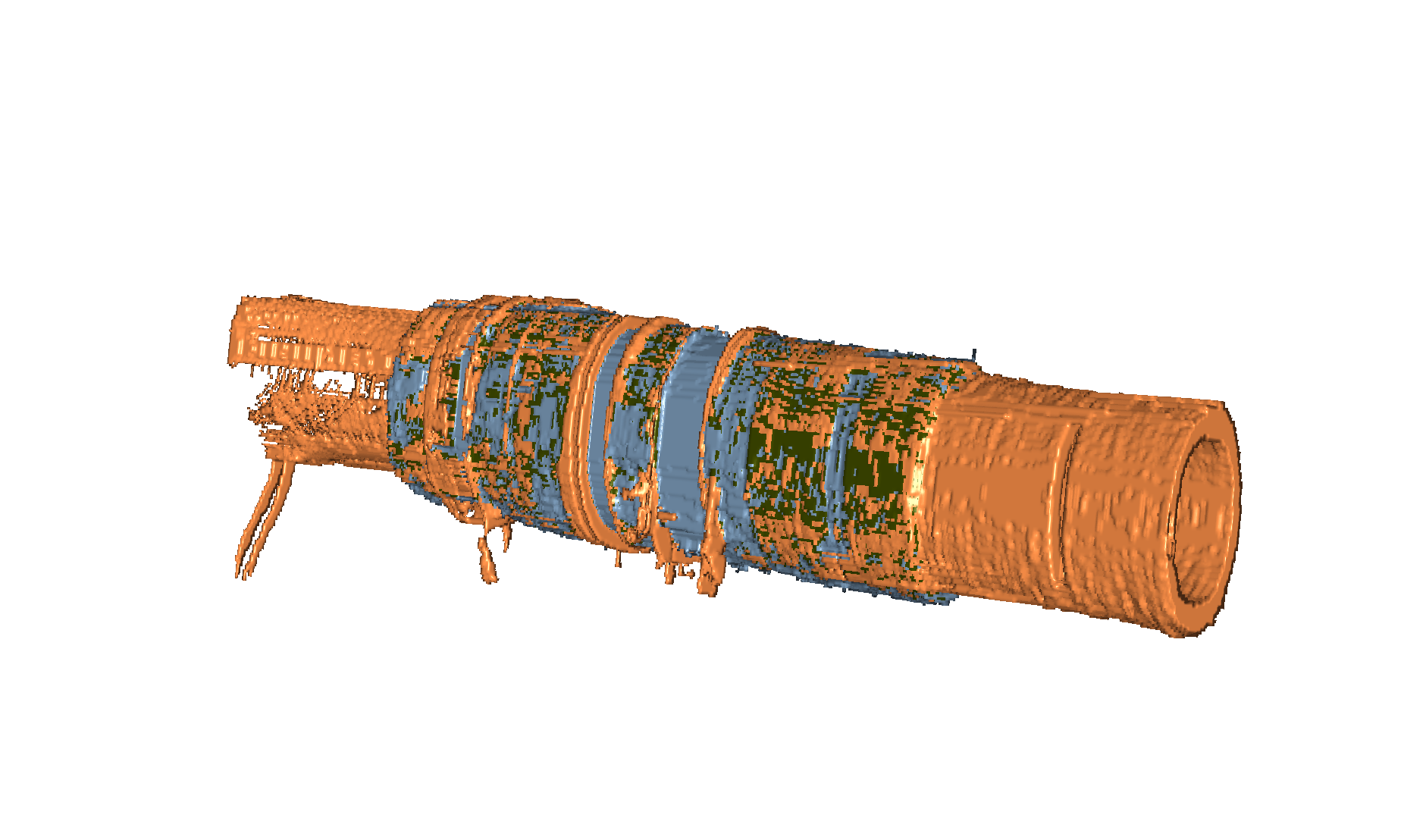}}
		 		~\subfloat[IoU~0.32]{\label{fig:result-big-michen-connectedComponent-reference124-proposed65}\includegraphics[width=0.25\columnwidth,keepaspectratio]{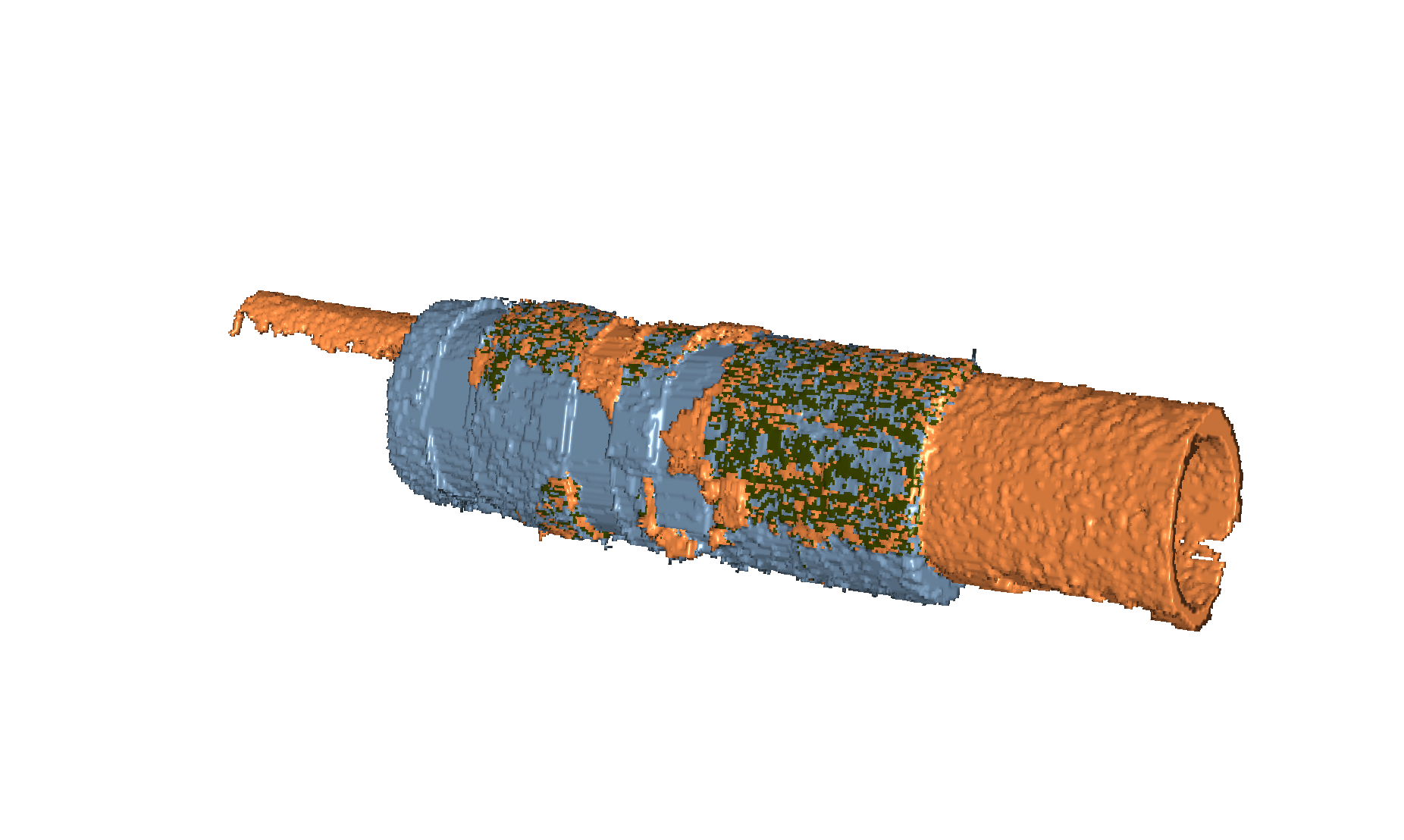}}					
		 		
		 		\caption{\label{fig:result-big} Renderings of the six largest voxel-count segments, in blue (\tikzBoxFalseNegative), against the proposed segmentation of \displayNameEngster~(Figures \labelcref{fig:result-big-engster-connectedComponent-reference31-proposed306,fig:result-big-engster-connectedComponent-reference30-proposed597,fig:result-big-engster-connectedComponent-reference24-proposed450,fig:result-big-engster-connectedComponent-reference23-proposed363,fig:result-big-engster-connectedComponent-reference25-proposed417,fig:result-big-engster-connectedComponent-reference124-proposed883}) and \displayNameMichen~(Figures \labelcref{fig:result-big-michen-connectedComponent-reference31-proposed1,fig:result-big-michen-connectedComponent-reference30-proposed214,fig:result-big-michen-connectedComponent-reference24-proposed208,fig:result-big-michen-connectedComponent-reference23-proposed99,fig:result-big-michen-connectedComponent-reference25-proposed225,fig:result-big-michen-connectedComponent-reference124-proposed65}), in orange (\tikzBoxFalsePositive). The overlap of these segmentations is shown in green (\tikzBoxTruePositive).}
		 	\end{figure}

		 	Figure \ref{fig:result-big} shows the six largest reference segments, such as metal sheets and two pipes connected by a hose. For each segment the best-matching predictions from \displayNameEngster\, and \displayNameMichen\, are shown. As both algorithms tend to under segment, there are large areas in multiple results which are only minimally covered by the proposed segmentation, for example Figures \ref{fig:result-big-engster-connectedComponent-reference25-proposed417} and \ref{fig:result-big-michen-connectedComponent-reference25-proposed225}. The complexity stems from two factors. Firstly, the correlation matrix's rules allow only a single proposed segment for each reference segment. Secondly, the algorithms are prone to under-segmentation, which can lead to situations where the appropriately matched proposed segment for one reference segment was already associated with a different reference segment.	
		 	
\begin{figure*}
				\centering
				\subfloat[Reconstruction \label{fig:result-slice-reconstruction}]{\includegraphics[width=0.4\textwidth, interpolate=false,keepaspectratio]{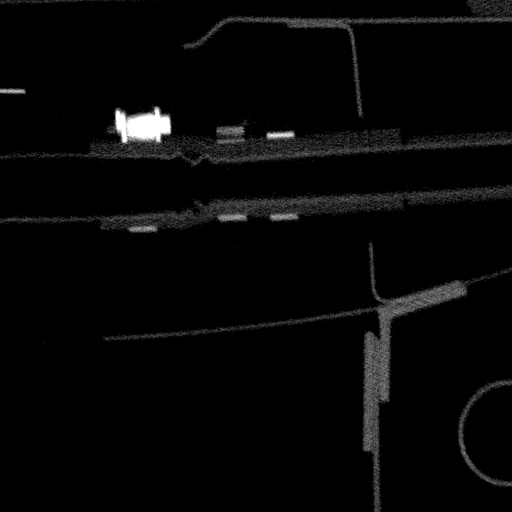}}	
				~\subfloat[Reference \label{fig:result-slice-reference}]
				{\includegraphics[width=0.4\textwidth, interpolate=false,keepaspectratio]{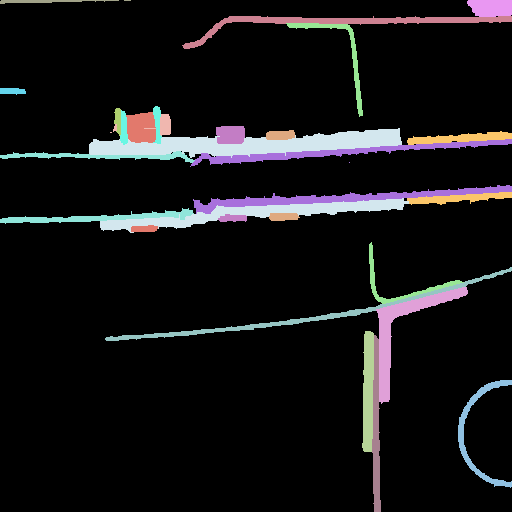}}
				
				\subfloat[\displayNameEngster \label{fig:result-slice-engster}]	
				{\includegraphics[width=0.4\textwidth, interpolate=false,keepaspectratio]{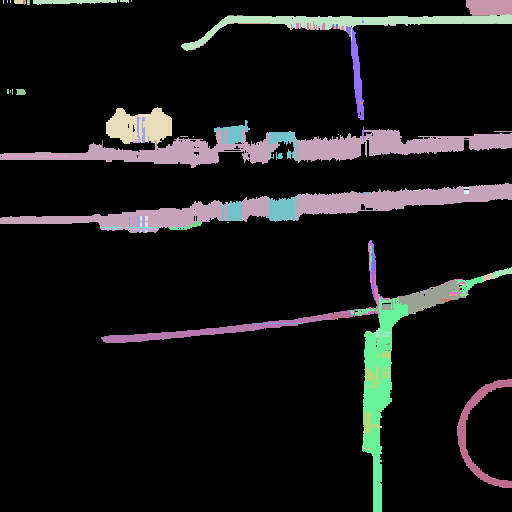}}	
				~\subfloat[\displayNameMichen \label{fig:result-slice-michen}]
				{\includegraphics[width=0.4\textwidth, interpolate=false,keepaspectratio]{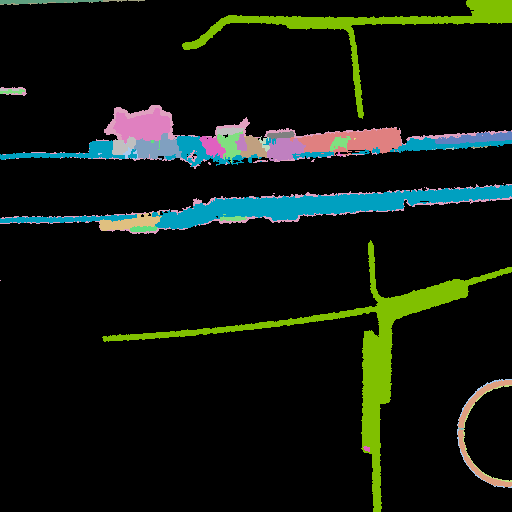}}	
				\caption{\label{fig:result-slice} An example slice of the $V_\text{test}$ sub-volume, showing the reconstruction (Figure \ref{fig:result-slice-reconstruction}), reference segmentation (Figure \ref{fig:result-slice-reference}), proposed segmentations from \displayNameEngster~(Figure \ref{fig:result-slice-engster}), and proposed segmentations from \displayNameMichen~(Figure \ref{fig:result-slice-michen}).}				
			\end{figure*}

\begin{figure}
				\centering
				\subfloat[Reconstruction \label{fig:result-bolts-reconstruction}]					
				{\includegraphics[width=0.25\columnwidth, interpolate=false,keepaspectratio]{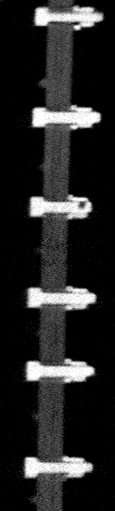}}
				~\subfloat[Reference \label{fig:result-bolts-reference}]					
				{\includegraphics[width=0.25\columnwidth, interpolate=false,keepaspectratio]{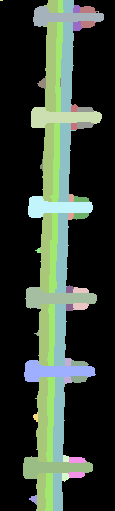}}	
				~\subfloat[\displayNameEngster \label{fig:result-bolts-engster}]
				{\includegraphics[width=0.25\columnwidth, interpolate=false,keepaspectratio]{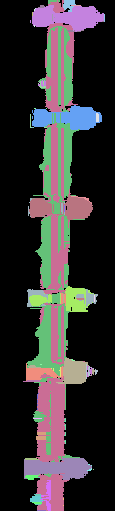}}	
				~\subfloat[\displayNameMichen \label{fig:result-bolts-michen}]
				{\includegraphics[width=0.25\columnwidth, interpolate=false,keepaspectratio]{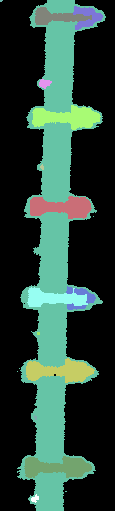}}
				\caption{\label{fig:result-bolts} 
				An example slice of the $V_\text{test}$ sub-volume showing a region with three metal sheets bolted and riveted together (Figure \ref{fig:result-bolts-reconstruction}), as well as the reference segmentation (Figure \ref{fig:result-bolts-reference}) and the proposed segmentations from \displayNameEngster~(Figure \ref{fig:result-bolts-engster}) and \displayNameMichen~(Figure \ref{fig:result-bolts-michen}).}
			\end{figure}

			For better visualization Figure \ref{fig:result-slice} depicts multiple versions of the same slice of $V_\text{test}$, with the reconstruction shown in Figure \ref{fig:result-slice-reconstruction}, reference segmentation in Figure \ref{fig:result-slice-reference}, and proposed segmentations from \displayNameEngster\,, and \displayNameMichen\, shown in Figures \ref{fig:result-slice-engster} and \ref{fig:result-slice-michen}, respectively. Figure \ref{fig:result-bolts} displays the same volumes but a different challenging region with three metal sheets bolted and riveted together.
			
			Comparing the two algorithms, it appears that the algorithm by \displayNameEngster\, can segment the large metal sheets quite well. However, typical errors of the \displayNameEngster\, algorithm include stripey artefacts arising from the merging of its intermediate 2D results, as seen in Figure \ref{fig:result-big-engster-connectedComponent-reference23-proposed363}. The matching algorithm also tends to miss boundaries between multiple full-contact metal sheets, particularly if they are of low contrast, as evidenced by the lower green segment in Figure \ref{fig:result-slice-engster} and the metal sheets in Figure \ref{fig:result-bolts-engster}, where 2D artefacts introduce new segments inside the metal sheets. It also tends to include entities in the vicinity of the current segment which can be partly corrected by the proposed connected components postprocessing, if the segments do not touch each other. However, for example Figures \ref{fig:result-big-engster-connectedComponent-reference24-proposed450} and \ref{fig:result-big-engster-connectedComponent-reference124-proposed883} demonstrate, that this is not always a viable solution, as many erroneously segmented segments are touching and are therefore difficult to separate.
			
			In comparison, under challenging circumstances or error conditions, the approach by \displayNameMichen\, tends to produce segmentations which are more irregular or patchy in appearance, as evidenced by the central region in Figure \ref{fig:result-slice-michen}. It also tends to combine metal sheets with low contrast, as observed in the large green segment at the top and bottom of the same figure. This combined segment is also depicted in Figure \ref{fig:result-big-michen-connectedComponent-reference31-proposed1}. As the algorithm uses classification between different classes to identify background--foreground, and foreground--background boundaries, a typical error case involves the creation of thin borders or wrapping skins around segments, as seen in the otherwise well segmented segment such as the pipe in the lower right corner of Figure \ref{fig:result-slice-michen} and around the segments in Figure \ref{fig:result-bolts-michen}. 
		
\begin{figure}
				\centering
					\subfloat[IoU 0.88]					
					{\includegraphics[width=0.33\columnwidth,keepaspectratio]{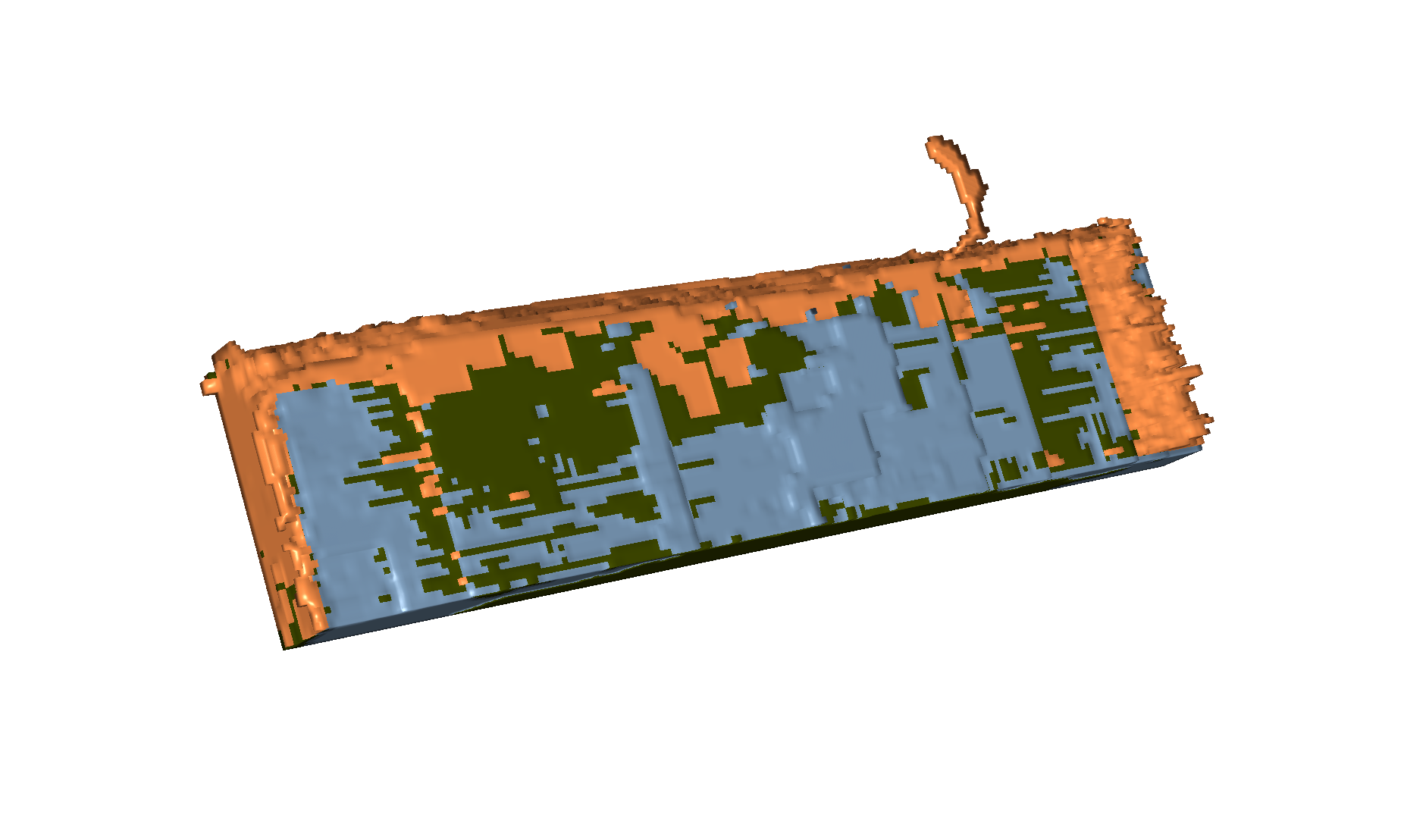}}
					~\subfloat[IoU 0.73]					
					{\includegraphics[width=0.33\columnwidth,keepaspectratio]{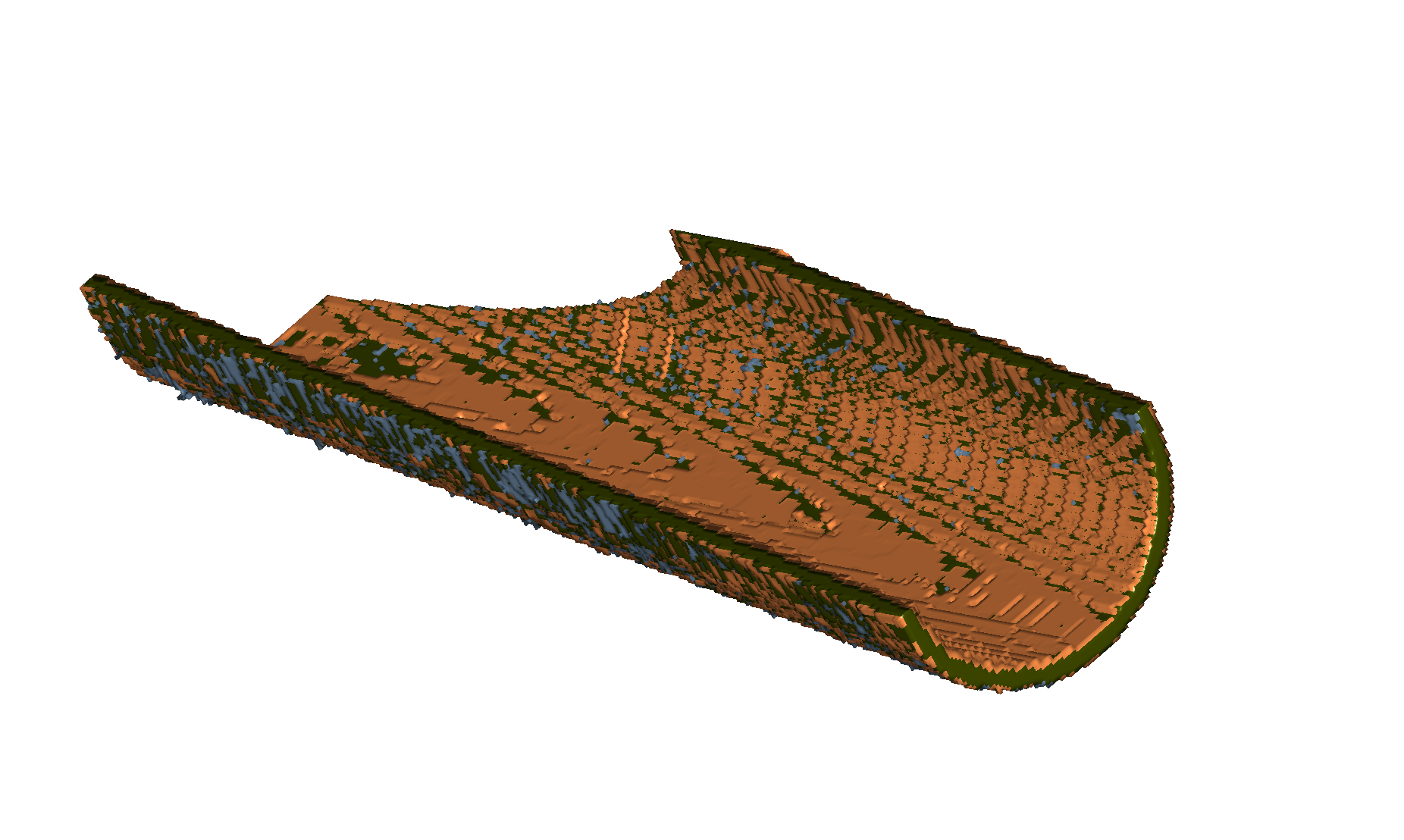}}
					~\subfloat[IoU 0.67]					
					{\includegraphics[width=0.33\columnwidth,keepaspectratio]{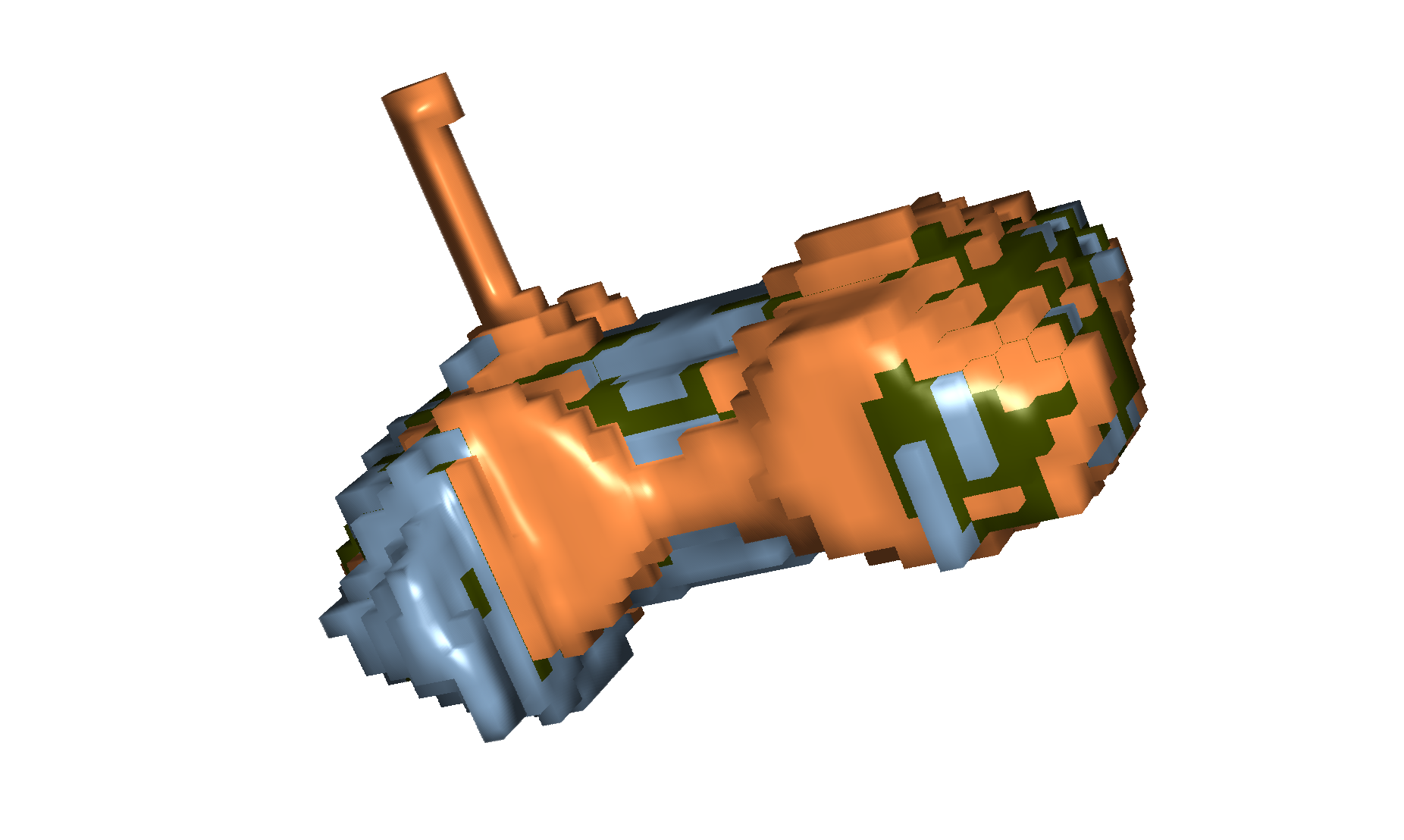}}				
					
					\subfloat[IoU 0.64]					
					{\includegraphics[width=0.33\columnwidth,keepaspectratio]{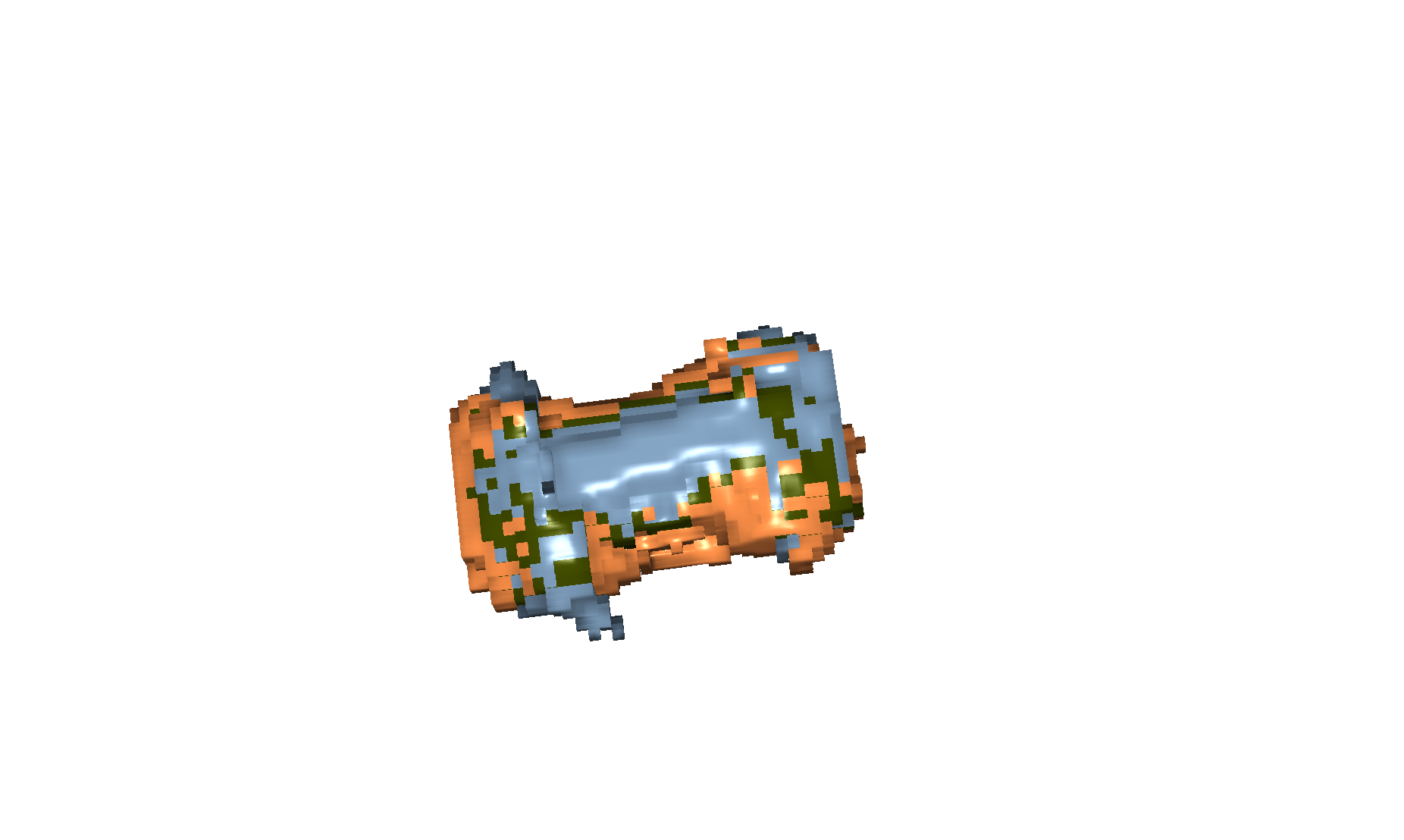}}	
					~\subfloat[IoU 0.62]					
					{\includegraphics[width=0.33\columnwidth,keepaspectratio]{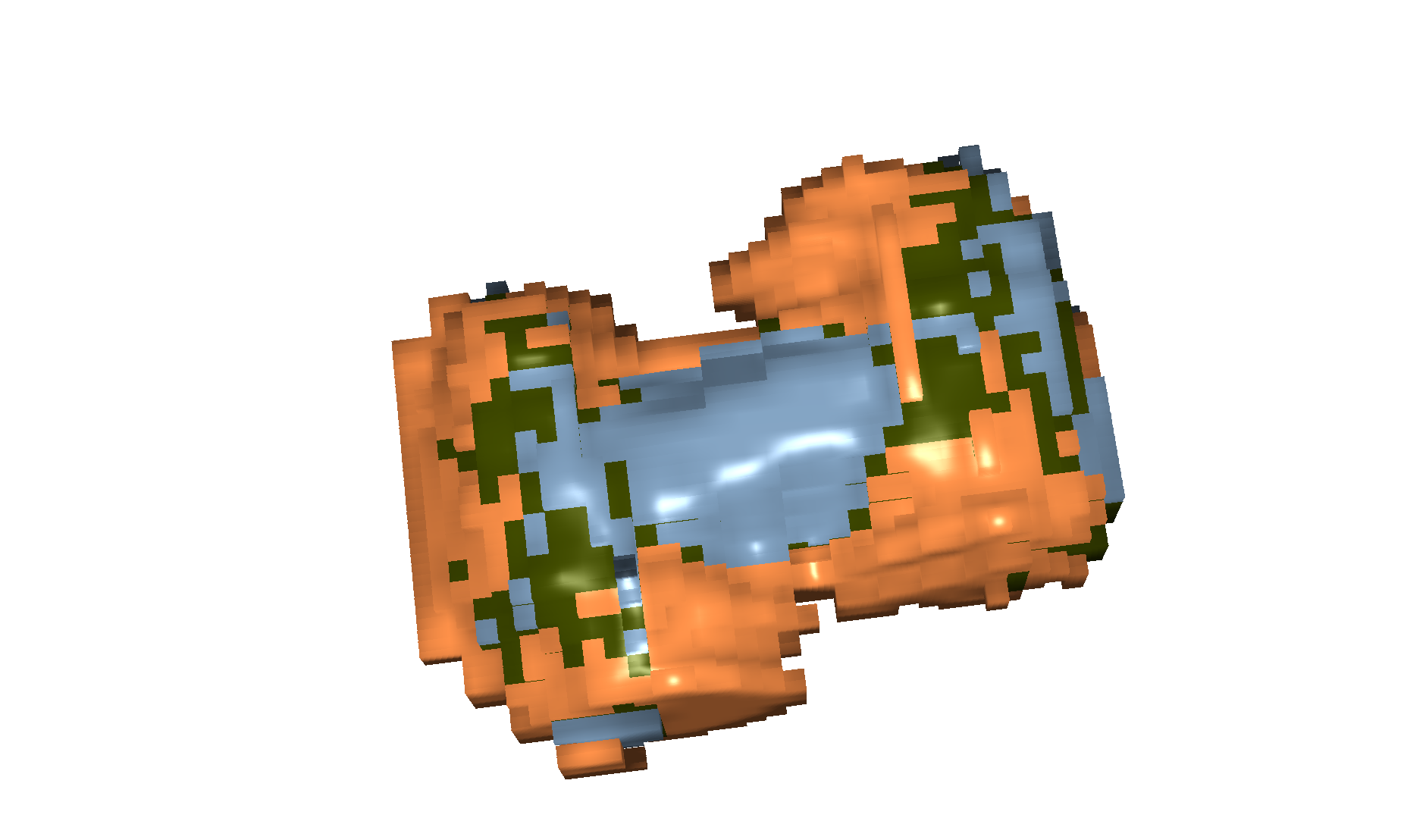}}	
					~\subfloat[IoU 0.61]					
					{\includegraphics[width=0.33\columnwidth,keepaspectratio]{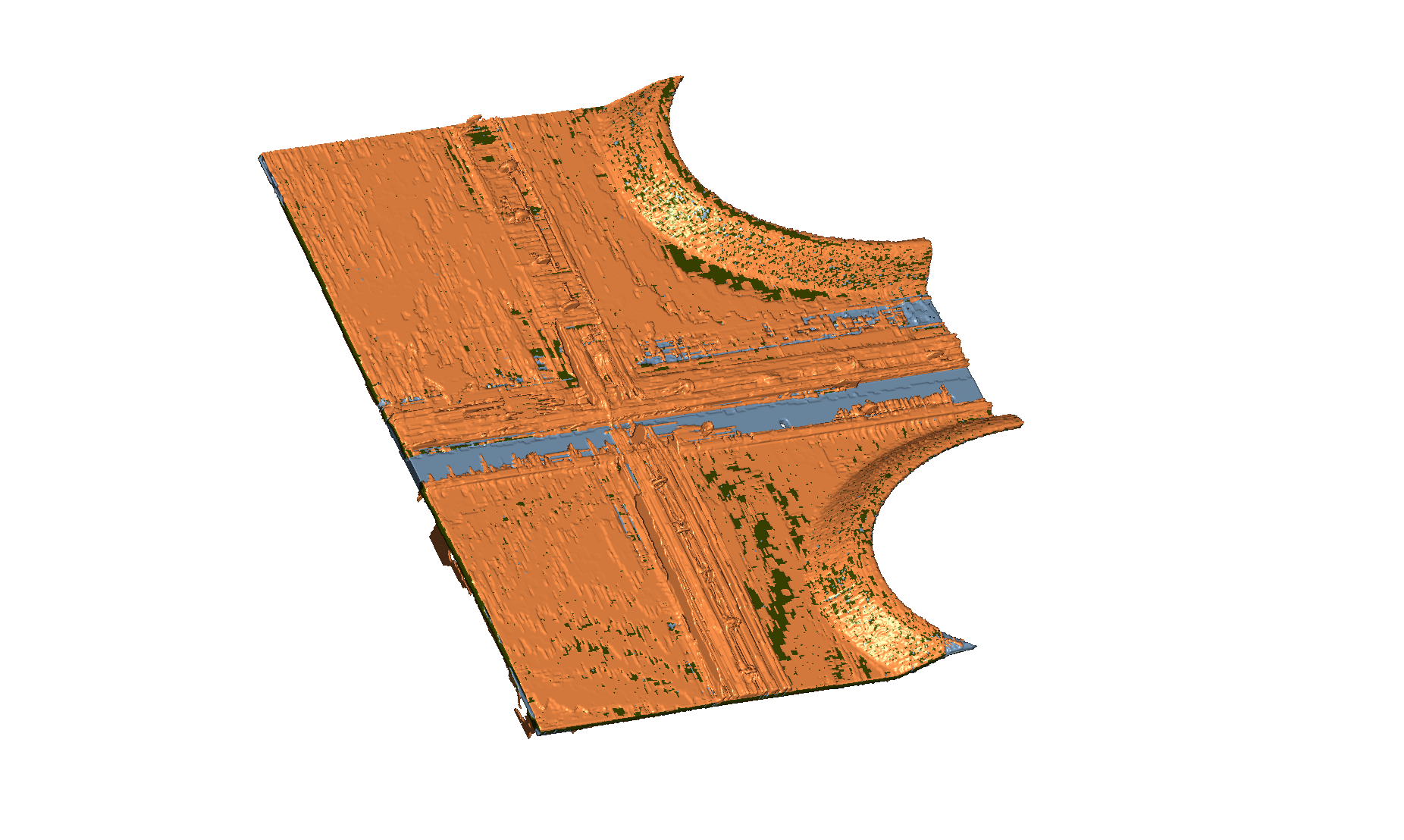}}				
				\caption{\label{fig:result-top-engster} Renderings of the six proposed sections by \displayNameEngster~which have the highest IoU with respect to the reference segments, in blue~( \tikzBoxFalseNegative). Proposed segments in orange~(\tikzBoxFalsePositive). The overlap of these segmentations is shown in green~(\tikzBoxTruePositive).}
			\end{figure}

\begin{figure}
				\centering
					\subfloat[IoU 0.85]					
					{\includegraphics[width=0.33\columnwidth,keepaspectratio]{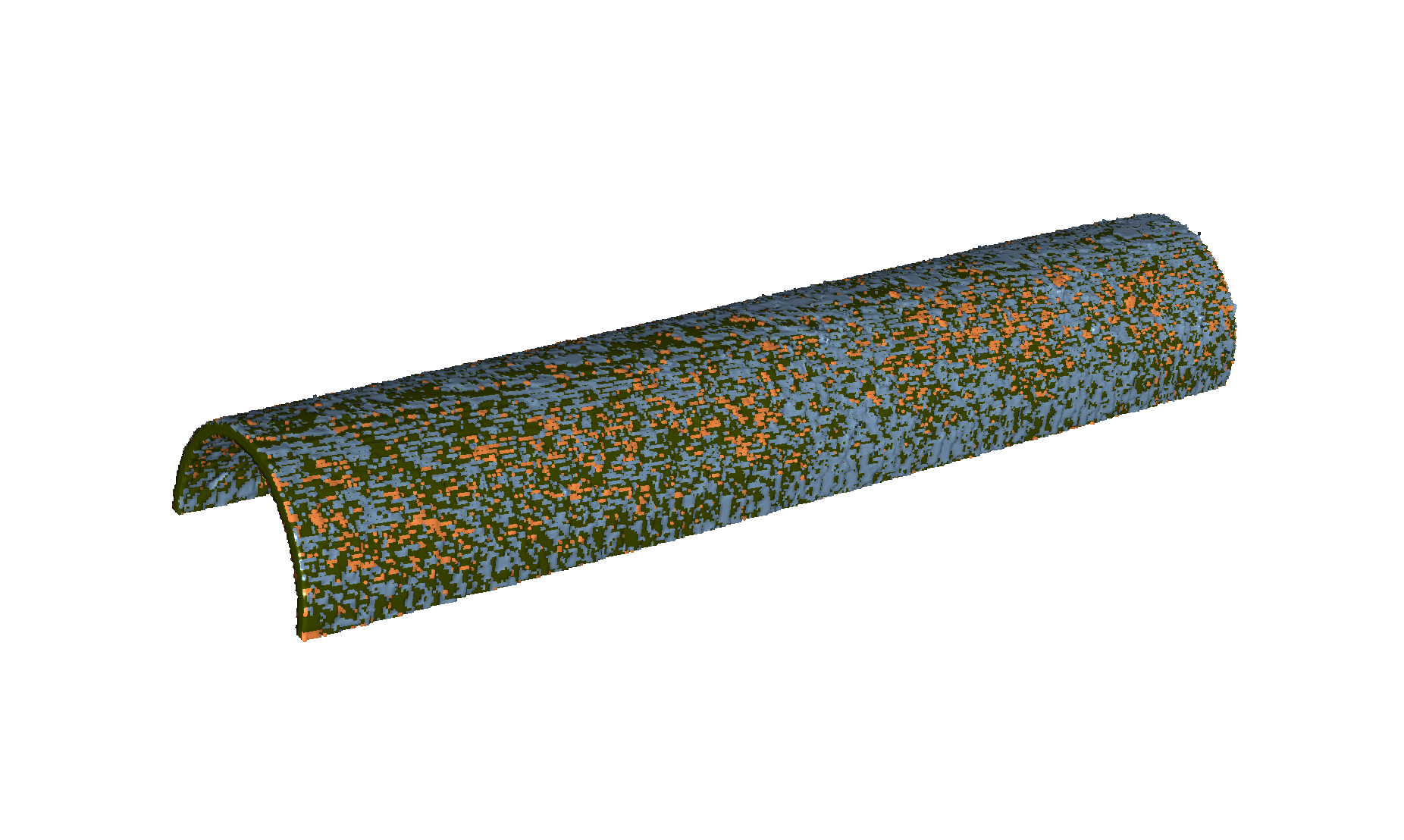}}
					~\subfloat[IoU 0.85]					
					{\includegraphics[width=0.33\columnwidth,keepaspectratio]{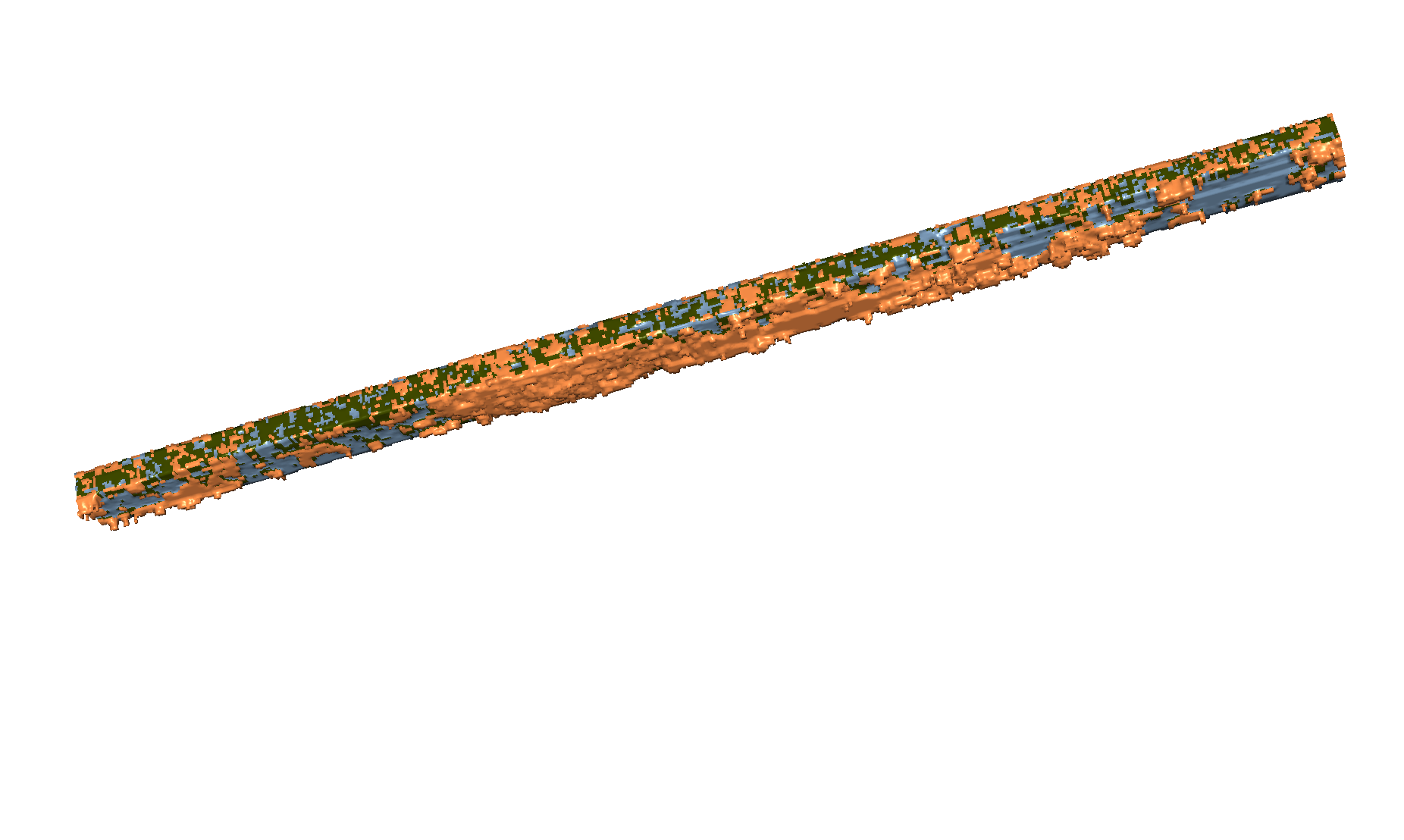}}
					~\subfloat[IoU 0.73]					
					{\includegraphics[width=0.33\columnwidth,keepaspectratio]{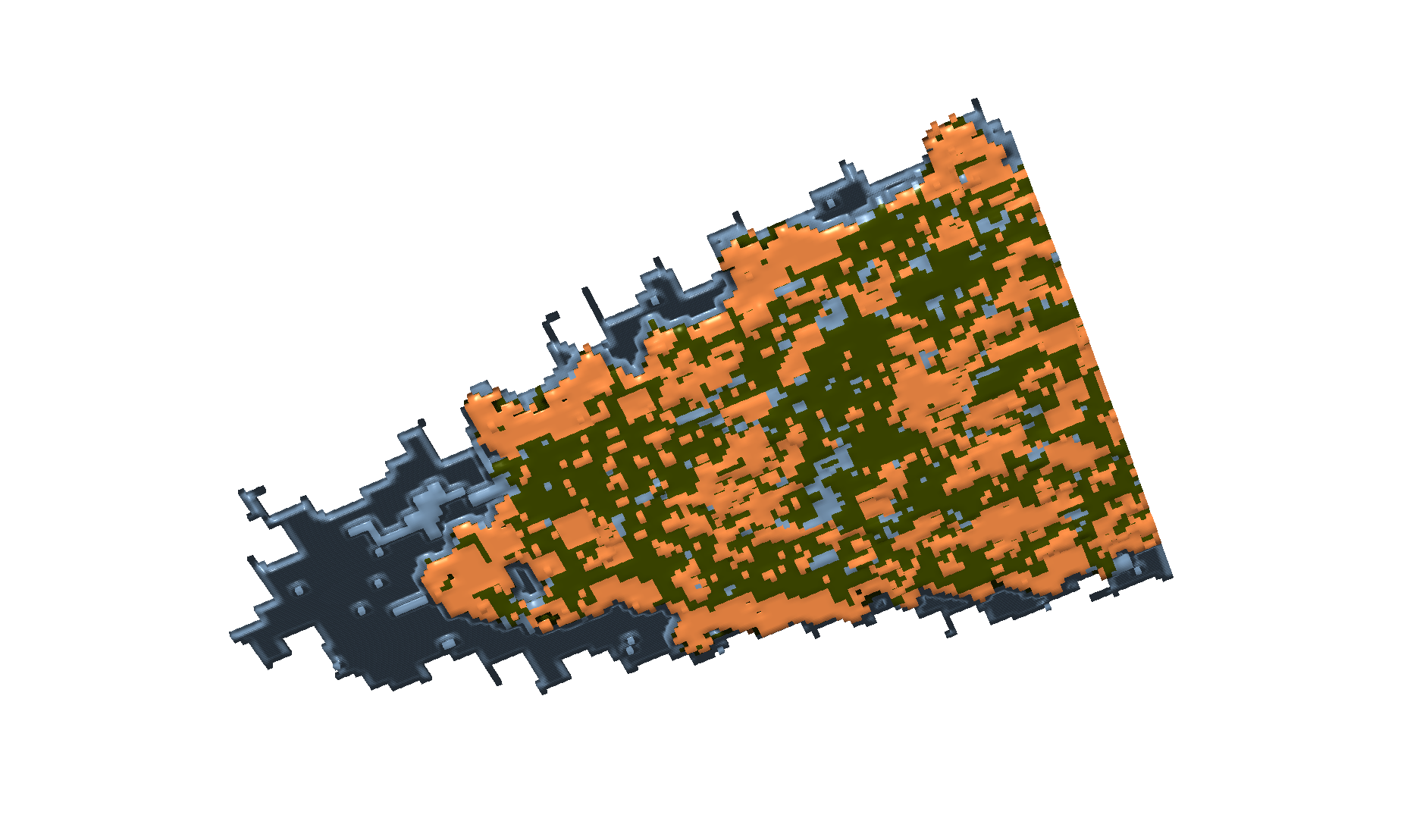}}
					
					~\subfloat[IoU 0.69]					
					{\includegraphics[width=0.33\columnwidth,keepaspectratio]{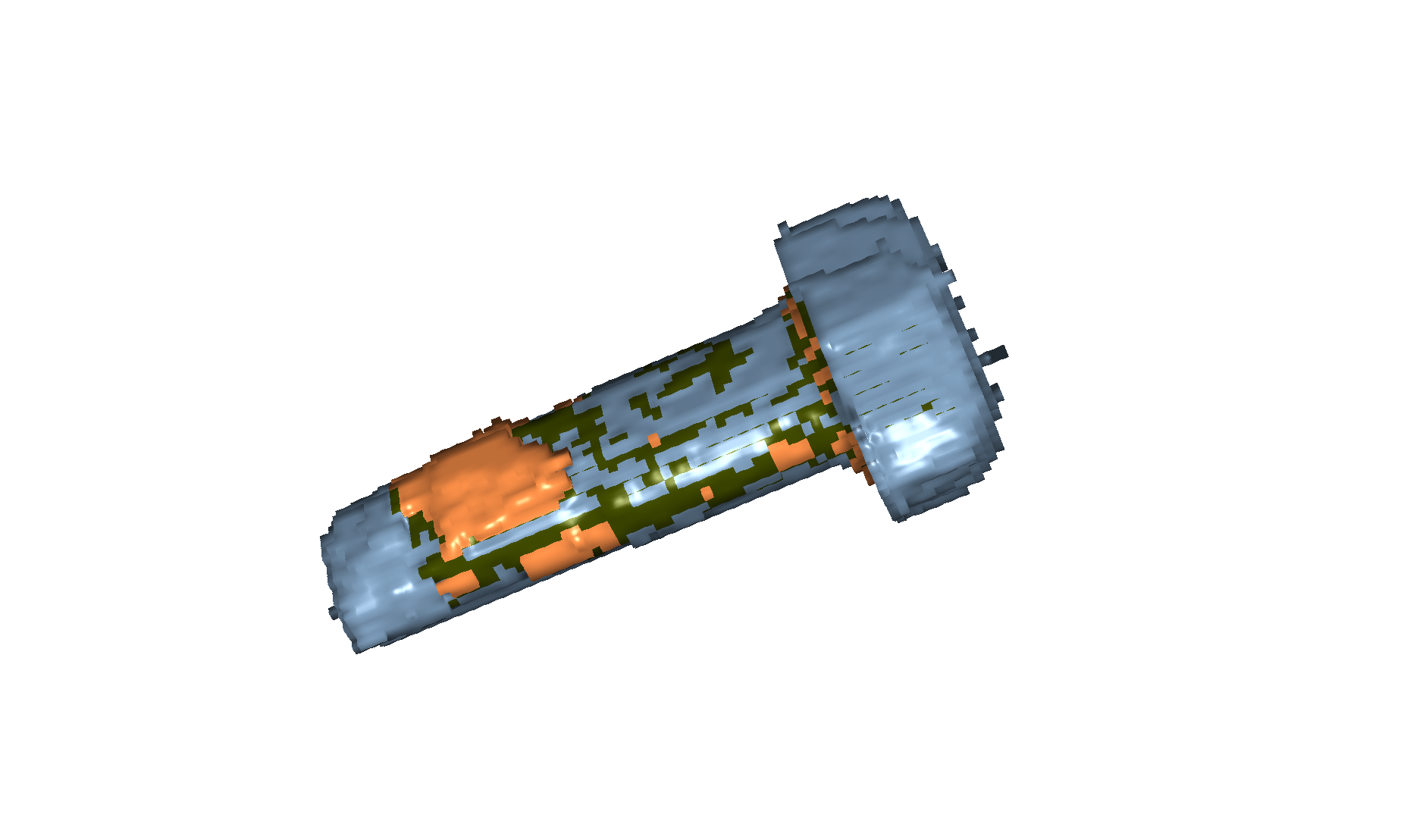}}	
					~\subfloat[IoU 0.69]					
					{\includegraphics[width=0.33\columnwidth,keepaspectratio]{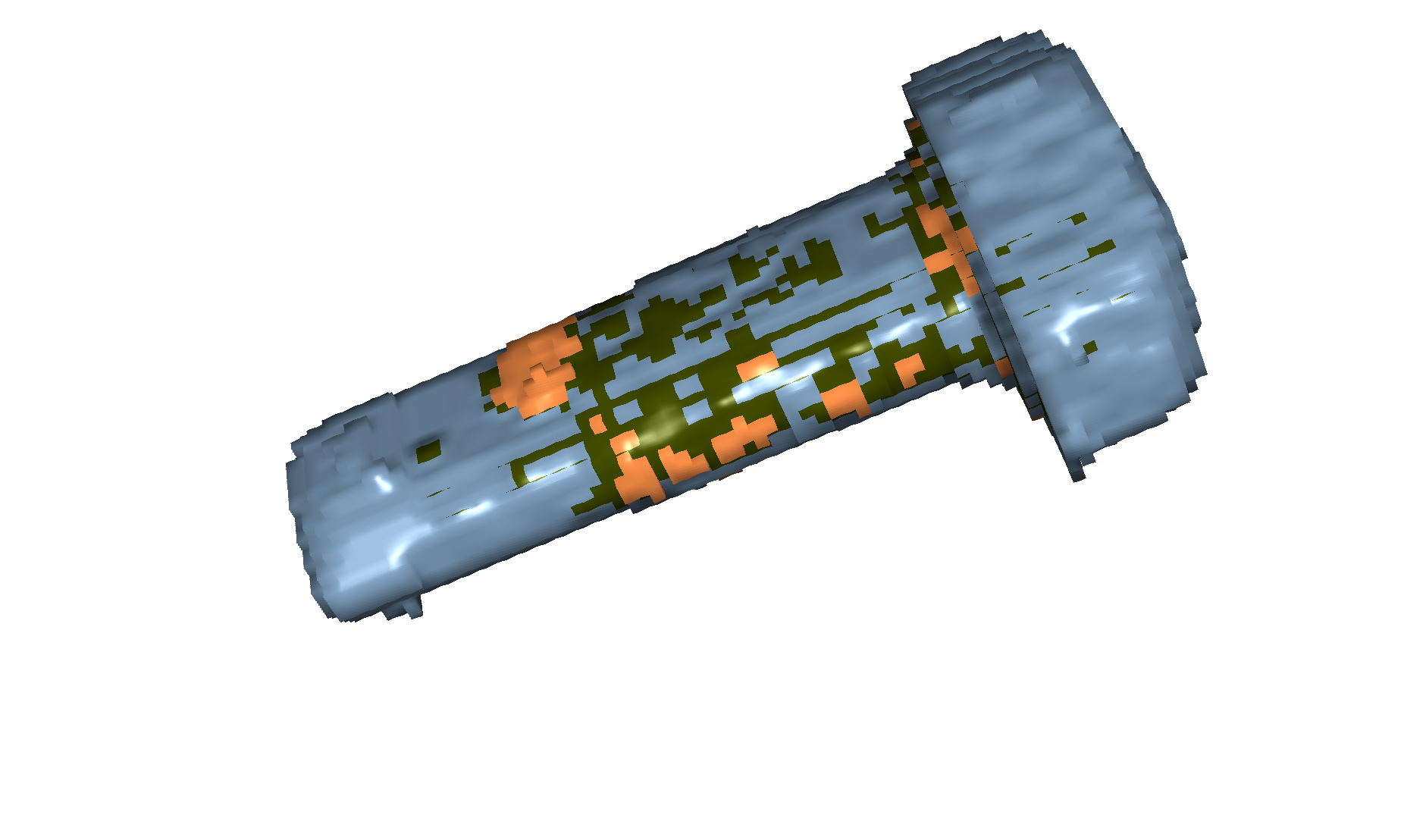}}	
					~\subfloat[IoU 0.54]					
					{\includegraphics[width=0.33\columnwidth,keepaspectratio]{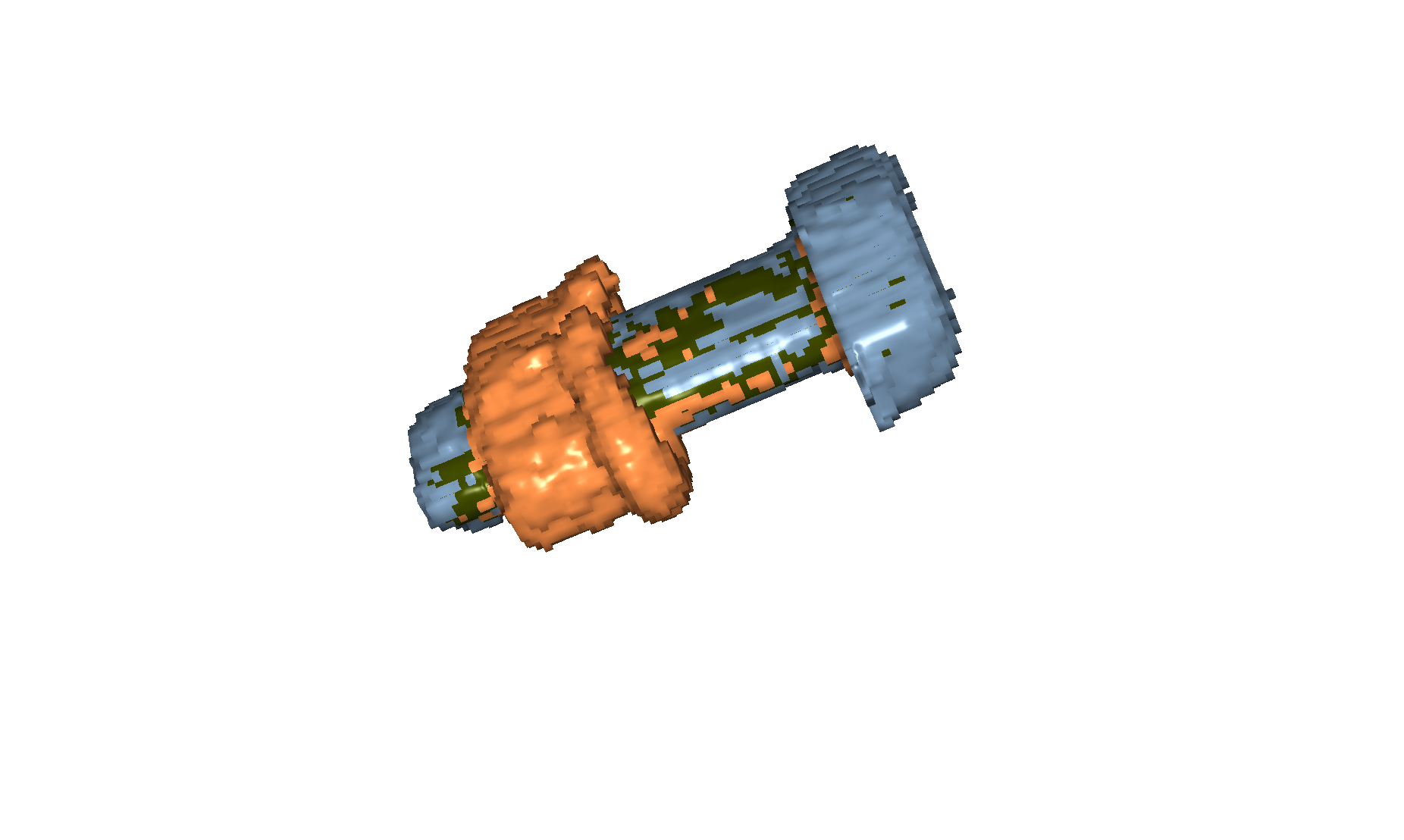}}	
				\caption{\label{fig:result-top-michen} Renderings of the six proposed sections by \displayNameMichen~ which have the highest IoU with respect to the reference segments, in blue~(\tikzBoxFalseNegative). Proposed segments in orange~(\tikzBoxFalsePositive). The overlap of these segmentations is shown in green~(\tikzBoxTruePositive).}
			\end{figure}

			Finally, Figures \ref{fig:result-top-engster} and \ref{fig:result-top-michen} display the six best segmented segments by IoU for \displayNameEngster\, and \displayNameMichen, respectively. It is evident, for instance, that among the top six detections by \displayNameEngster, there are three large objects (two plates and one pipe) as well as three small rivets. By comparison, the top detections by \displayNameMichen\, consist of one large segment, specifically a pipe, along with five smaller segments, namely a rod, a sheet of metal, and three screws. This observation reinforces the notion that \displayNameEngster\, excels in identifying larger objects, while \displayNameMichen, demonstrates proficiency in detecting smaller items.
	
	\section{Discussion}

		The challenge described above focused on the instance segmentation of a specific NDT data-set obtained from a historical airplane which was created using XXL-CT imaging. Originally eleven participants registered for the challenge, from whom two finally submitted their segmentation proposals. Five participants withdrew from the challenge providing feedback and information why they did not submit a segmentation proposal for the challenge data-set. Despite multiple attempts to contact the remaining four participants to either finish the challenge or officially withdraw, they did not react to our outreach or were unable to be contacted. Ultimately, we enforced the deadline at end of August 2023 to conclude the challenge.
		
		The segmentation proposals submitted by \displayNameEngster\, and \displayNameMichen\, utilized different instance segmentation approaches. \displayNameEngster\, employed a 2D image transformer followed by 3D matching, while \displayNameMichen\, utilized a 3D U-Net for segmenting foreground, background, and border classes.	In some instances, one algorithm slightly outperformed the other, such as \displayNameEngster\, in segmenting larger metal sheets and \displayNameMichen\, in segmenting smaller entities.
		
		Based on the findings of our analysis, we observed promising results in terms of 3D instance segmentations, compared to the manual annotated reference data. However, the development of robust methods for instance segmentation of XXL-CT data still remains a challenge, particularly due to the limited availability of annotated reference data-sets and the limited data quality.
		
		Carrying out of the reported XXL-CT segmentation challenge has certainly raised the awareness within the NDT community regarding novel imaging modalities and the necessity to explore innovative methods for related segmentation tasks. Specifically, XXL-CT represents a cutting-edge imaging modality which enables the volumetric imaging of large-scale objects across various domains. The challenge has unraveled the need for further research and the development of new methodologies to achieve comprehensive instance segmentation in this context, as conventional segmentation approaches are inadequate.
	
	\section{Acknowledgements} 
	
		This work was supported by the Bavarian Ministry of Economic Affairs, Regional Development and Energy through the Center for Analytics Data Applications (ADA-Center) within the framework of ,,BAYERN DIGITAL II`` (20-3410-2-9-8).
	
		We want to thank the following colleagues for the manual annotation: Verena Malowaniec, Laura Heidner, and Kseniia Dudchenko.
	
	\section*{Author contributions statement} 

		\begin{description}
			\item [{RG}] – organized, supervised and evaluated the challenge, he also supervised the testing data annotation, designed and implemented the used evaluation metrics. He drafted and wrote the manuscript including the graphics.
			\item [{JCE}] – together with NB and MS participated in the challenge, he coordinated the contribution of \displayNameEngster. Further, he implemented, trained, and tested the \displayNameEngster~approach and wrote Section \ref{sec:results-engster}.
			\item [{MM}] – participated in the challenge, he implemented trained and tested the \displayNameMichen~approach  and wrote Section \ref{sec:results-michen} about the \displayNameMichen~approach.
			\item [{NB}] – together with JCE and MS participated in the challenge, she implemented trained and tested the \displayNameEngster~approach and contributed to Section \ref{sec:results-engster} about the \displayNameEngster~approach.
			\item [{MS}] – together with JCE and NB participated in the challenge, he implemented trained and tested the \displayNameEngster~approach and contributed to Section \ref{sec:results-engster} about the \displayNameEngster~approach.
			\item [{TW}] - together with RG – provided the original idea if the challenge, planned, concepted, and supported the challenge, wrote the introduction and the setting, and did the final proofreading and editing.
		\end{description}	

		All authors reviewed the manuscript.
	
	\section*{Competing interests} 
		The authors declare that they have no competing financial and/or non-financial interests in relation to the work described.
		
	\bibliographystyle{plainurl}
	\bibliography{challenge}	
	
\end{document}